\documentclass[letterpaper]{article} 
\usepackage[preprint]{aaai2027}
\usepackage[hyphens]{url}  
\usepackage{graphicx} 
\usepackage{natbib}  
\usepackage{caption} 
\usepackage{booktabs}
\usepackage{amsmath}
\usepackage{amssymb}
\usepackage{multirow}
\usepackage{tikz}

\usetikzlibrary{arrows.meta, positioning, calc}

\title{Bidirectional Preference Synthesis: Learning Prompt-Conditioned Preferences from Boundary Failures}

\author{
    Junbo Wang\textsuperscript{\rm 1,2},
    Lidong Lu\textsuperscript{\rm 2},
    Zhuoqun Li\textsuperscript{\rm 1},
    Guiping Jiang\textsuperscript{\rm 1},
    Xiangyu Wu\textsuperscript{\rm 1},
    Tinghai Zhang\textsuperscript{\rm 1},
    Tong Lu\textsuperscript{\rm 2}
}
\affiliations{
    \textsuperscript{\rm 1}Kuaishou Technology\\
    \textsuperscript{\rm 2}Nanjing University
}

\begin{document}
\maketitle

\begin{abstract}
Correction-based offline preference pipelines commonly treat model failures only as rejected responses under the original prompt. This supervision is incomplete for \emph{boundary failures}: responses that violate the given instruction yet coherently satisfy a nearby intent or constraint setting. We introduce \textbf{Bidirectional Preference Synthesis} (BPS), a data-construction method for standard DPO that makes this missing prompt dependence explicit. For each validated boundary failure, BPS keeps the conventional forward pair under the original prompt and adds a reverse pair under a synthesized \emph{achieved prompt}, so the same response is rejected where it is wrong and chosen where it is right, without changing the DPO objective, training a reward model, or requiring online sampling. On Qwen3-4B-Instruct-2507, BPS preserves original-side pairwise ranking while raising achieved-side ranking accuracy from 6.8\% to 62.3\% on held-out crossed anchors, with a similar shift under a Kimi-K2.6 cross-teacher probe. A blind human audit supports the intended reverse preference direction, and downstream evaluations show the clearest separation from Forward-DPO in multilingual multi-turn instruction following, with consistent capability-retention patterns on agentic/tool-use and code checks.
\end{abstract}

\section{Introduction}

A rejected response in preference data is not always bad in itself. Consider a prompt asking for a Chinese answer in exactly two bullet points, with a required keyword and separator. A model may produce a fluent, factually correct Chinese answer, but use three bullets or omit the separator. Under the original prompt, this response should lose to a correction. Yet the same response could be better under a prompt whose constraints match what the model produced.

We call such near misses \textbf{boundary failures}: outputs that violate the given prompt but coherently satisfy a related intent or constraint setting. They arise naturally in instruction-following tasks with verifiable constraints~\cite{zhou2023instruction,pyatkin2025generalizingverifiableinstructionfollowing,jiang2024followbench}, where a response can solve the task while missing a format, lexical, language, or structural requirement. The same structure appears beyond surface constraints: a tool argument can be wrong for one request but correct for a related request, and a multilingual pattern can be inappropriate in one turn but useful in another~\cite{he2024multi,qin2023toolllmfacilitatinglargelanguage,yao2024taubenchbenchmarktoolagentuserinteraction}. Boundary failures are therefore not merely negative examples; they expose where preference should change with the prompt.

Offline preference optimization methods such as DPO~\cite{rafailov2023direct} and its variants~\cite{azar2024general,ethayarajh2024ktomodelalignmentprospect,meng2024simpo,hong2024orpo} are widely used because they avoid reward-model RLHF pipelines~\cite{christiano2017deep,stiennon2020learning,ouyang2022training} and online PPO-style rollouts. Their effect, however, is determined by the preference pairs they receive. A typical pipeline samples failed rollouts, obtains corrected responses through rejection sampling~\cite{touvron2023llama2openfoundation} or self-correction~\cite{zelikman2022star,madaan2023self}, and trains on pairs where the correction is preferred under the original prompt. We call this \emph{forward-only} supervision because each failure is used only as rejected under its original prompt. This is the right signal when the rejected response is incoherent, unsafe, or factually wrong. For a boundary failure, it records only that the response lost under $P_\text{orig}$, not whether the rejection is prompt-specific.

This missing context can matter even though DPO scores responses conditionally on the prompt. Because prompts share model parameters, repeatedly observing a coherent response mode only as rejected may influence its relative score in other contexts where that mode remains useful. We therefore study boundary failures as a data-construction problem: preference data can encode not only which response is wrong, but also where it would be right.

\begin{figure}[t]
\centering
\includegraphics[width=\columnwidth]{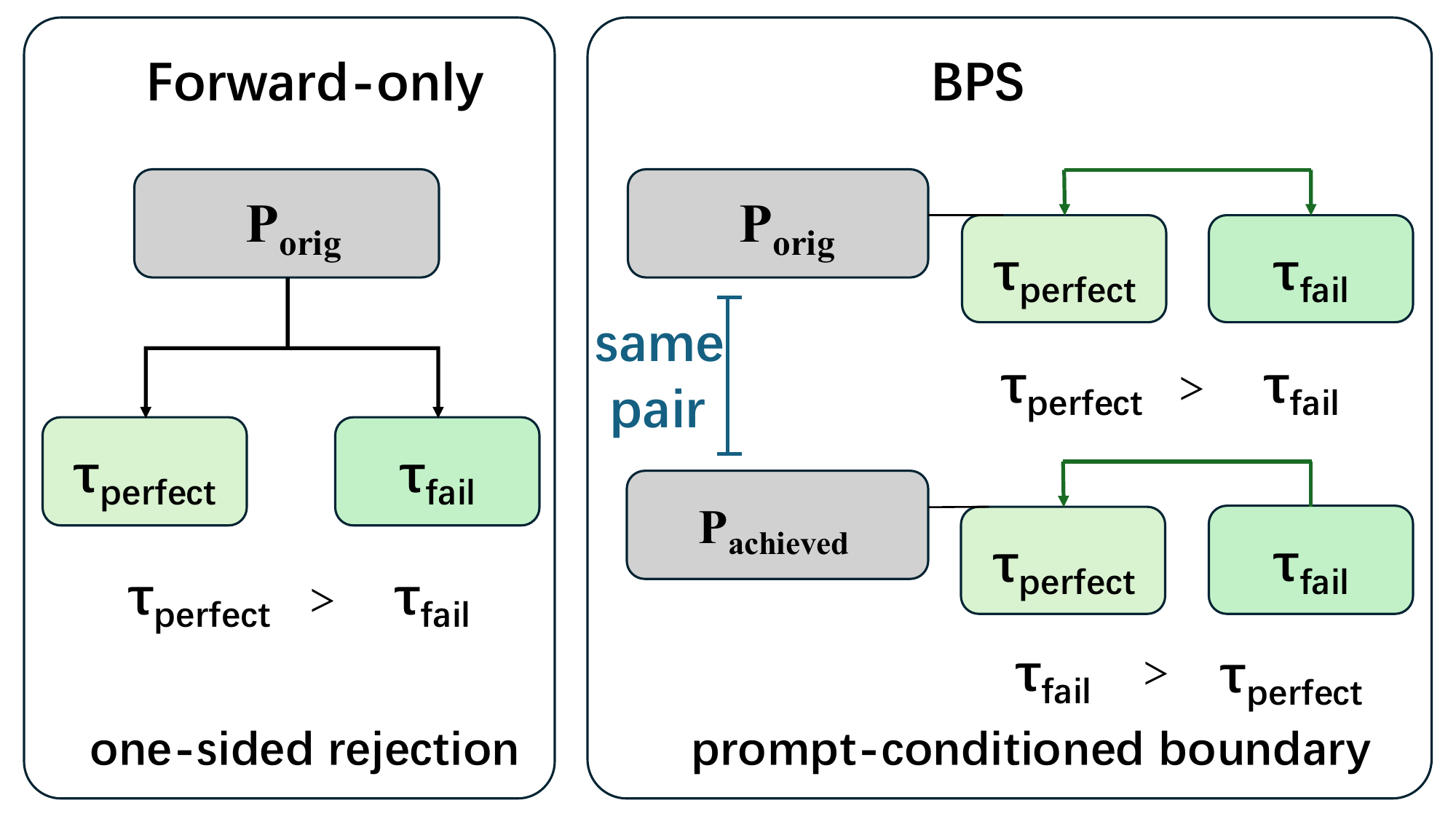}
\caption{Forward-only DPO data rejects $\tau_\text{fail}$ only under $P_\text{orig}$. BPS adds an achieved-prompt reverse pair, so the same response pair teaches a prompt-conditioned preference boundary.}
\label{fig:concept}
\end{figure}

We propose \textbf{Bidirectional Preference Synthesis} (BPS), an offline construction method for standard DPO. For each validated boundary failure, BPS keeps the conventional forward pair under the original prompt, $P_\text{orig}: \tau_\text{perfect} \succ \tau_\text{fail}$, and adds a reverse pair under a synthesized \emph{achieved prompt}, $P_\text{achieved}: \tau_\text{fail} \succ \tau_\text{perfect}$. The achieved prompt describes what the failed response satisfies. Together, the pairs form a crossed anchor: $\tau_\text{fail}$ is rejected where it is wrong and chosen where it is right. Figure~\ref{fig:concept} contrasts this with forward-only data. BPS changes only the preference data; it introduces no new loss, trained reward model, or online sampling.

We evaluate BPS on Qwen3-4B-Instruct-2507. Held-out crossed-anchor probes show that BPS preserves original-side rankings while substantially improving achieved-side rankings under both Gemini and Kimi-K2.6 construction pipelines. A blind human audit supports the intended reverse preference direction and finds little response leakage. Downstream evaluations show the clearest separation in multilingual multi-turn instruction following, together with favorable capability-retention patterns on agentic/tool-use and code checks. A separate function-calling instantiation extends the same construction to structured action spaces.

Our contributions are:
\begin{enumerate}
    \item We identify boundary failures as a preference-data construction problem: a rejected response may be useful under a related intent, so forward-only supervision can conflate prompt-specific rejection with response-intrinsic undesirability.
    \item We introduce Bidirectional Preference Synthesis, which augments validated boundary failures with achieved-prompt reverse pairs while leaving the DPO objective unchanged.
    \item We provide evidence that semantically matched reverse anchors counteract cross-prompt rejection patterns and improve cross-context capability retention, including multilingual interaction and function calling.
\end{enumerate}

\section{Related Work}

\paragraph{Preference optimization and DPO failure modes.}
Online RLHF methods train reward models from preference comparisons and optimize policies with RL-style updates~\cite{christiano2017deep,stiennon2020learning,ouyang2022training}, but require rollouts, reward queries, and stability tuning. Direct or offline objectives such as SLiC-HF, RRHF, DPO, IPO-style objectives, KTO, SimPO, and ORPO~\cite{zhao2023slichfsequencelikelihoodcalibration,yuan2023rrhfrankresponsesalign,rafailov2023direct,azar2024general,ethayarajh2024ktomodelalignmentprospect,meng2024simpo,hong2024orpo} avoid online RL, making data construction central: behavior depends on which responses are rejected under which prompts. Prior work shows that DPO can reduce loss by displacing likelihood away from rejected responses rather than learning calibrated preferences~\cite{razin2025unintentional}; DPO-positive/Smaug addresses related objective-level issues~\cite{pal2024smaug}. BPS is complementary: it keeps standard DPO but changes the data semantics by asking when a rejected response should also be positive.

\paragraph{Reverse and input-side preference construction.}
RPO constructs reverse preference pairs by modifying instruction constraints so that an imperfect response becomes preferred on a related task~\cite{huang2025reverse}. IOPO evaluates the same output under different inputs to make preference learning more input-conditioned~\cite{zhang2025iopo}. BPS is closest to this line, but ties each reverse pair to an observed boundary failure: it starts from a student response that failed $P_\text{orig}$, synthesizes a standalone prompt for what the response achieved, and keeps the reverse label only after validation. Unlike mechanical constraint reversal, BPS keeps the forward and achieved-prompt reverse pairs as a crossed anchor over the same responses. We use an RPO-style construction as a strong baseline to isolate the value of semantically matched achieved prompts.

\paragraph{Data construction, self-correction, and hindsight relabeling.}
Offline pipelines use filtering, teacher feedback, rejection sampling~\cite{touvron2023llama2openfoundation}, STaR~\cite{zelikman2022star}, Self-Refine~\cite{madaan2023self}, AI feedback~\cite{lee2024rlaifvsrlhfscaling}, or self-rewarding methods to produce improved responses. BPS uses this correction signal but also asks whether a failure reveals a nearby instruction it already satisfies. This connects to hindsight experience replay~\cite{andrychowicz2017hindsight} and HIR~\cite{zhang2023wisdom}, but BPS does not replace failure with success: it preserves rejection under the original prompt and preference only under a validated achieved prompt.

\paragraph{Instruction following and tool use.}
Instruction tuning and synthetic instruction generation show that task specifications shape model behavior~\cite{wei2022finetunedlanguagemodelszeroshot,wang2022super,wang2023self}. Fine-grained instruction-following benchmarks~\cite{zhou2023instruction,pyatkin2025generalizingverifiableinstructionfollowing,jiang2024followbench} expose partial success over language, format, lexical, and task-content constraints, while multilingual multi-turn benchmarks~\cite{he2024multi} test cross-turn stability. Tool-use and agentic methods and benchmarks~\cite{schick2023toolformer,patil2024gorilla,yao2024taubenchbenchmarktoolagentuserinteraction,qin2023toolllmfacilitatinglargelanguage,patil2025berkeley} expose the same structure in action spaces, where an argument can be wrong for one request but correct for a nearby request. These settings naturally test prompt-conditioned preference learning.

\begin{figure*}[t]
\centering
\includegraphics[width=0.95\textwidth]{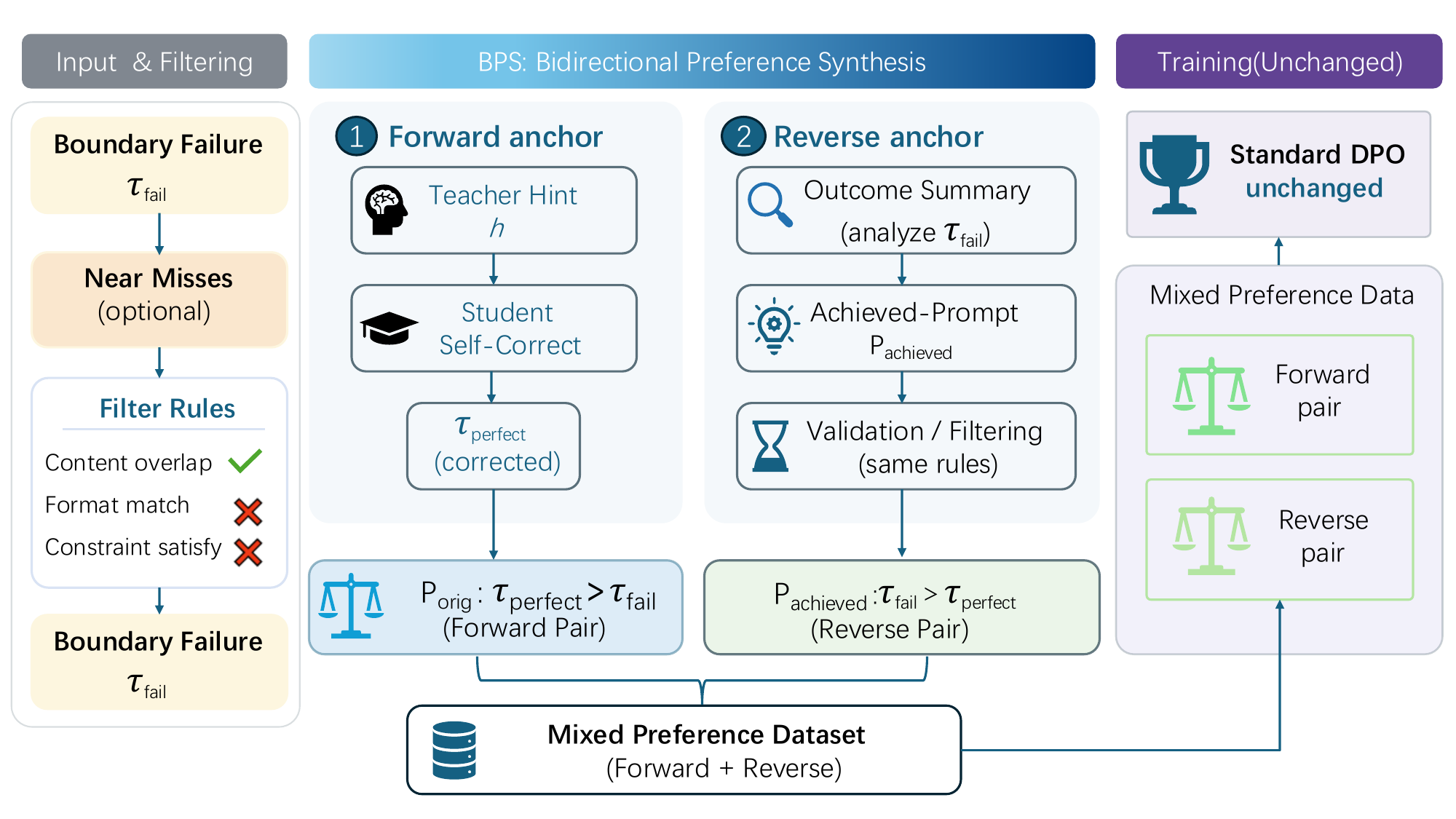}
\caption{BPS pipeline. Each validated boundary failure yields a forward correction pair under $P_\text{orig}$ and, when validation succeeds, a reverse pair under $P_\text{achieved}$.}
\label{fig:pipeline}
\end{figure*}

\section{Method}

\subsection{Problem Formulation}

BPS constructs preference pairs from model-generated near misses. Figure~\ref{fig:pipeline} summarizes the construction pipeline. It uses the student $\pi_{\theta_0}$ to generate failures and self-corrections, and a teacher $\pi_T$ only for offline diagnosis, correction hints, achieved-prompt synthesis, and validation. Given seed prompts $\mathcal{D}=\{P_i\}_{i=1}^N$ with verifiable atomic constraints $C(P_i)=\{c_1,\ldots,c_m\}$, we sample $K$ rollouts per prompt and define the \emph{constraint pass ratio}
\begin{equation}
r(P, \tau) = \frac{|\{c_k \in C(P) : \tau \text{ satisfies } c_k\}|}{|C(P)|}.
\end{equation}
Responses with $r=1$ need no correction, and responses with $r=0$ provide little evidence of a nearby achieved intent. BPS focuses on partial successes, $0<r<1$, especially near misses with $r\geq0.5$. A candidate receives a reverse anchor only after passing the eligibility and validation filters below.

\subsection{Forward Execution Anchor}

The forward anchor records the ordinary correction signal while keeping the corrected response student-generated. For each candidate boundary failure $(P_\text{orig}, \tau_\text{fail})$, the teacher produces a structured \emph{correction hint} $h$ rather than a replacement answer, and the student generates a correction conditioned on the original prompt, the failed response, and the hint:
\begin{equation}
\tau_\text{perfect} \sim \pi_{\theta_0}(\cdot \mid P_\text{orig},\ \tau_\text{fail},\ h)
\end{equation}
During this retry, $\tau_\text{fail}$ is evidence for repair, not an instruction source. A correction is accepted only if it is non-degenerate and achieves $r(P_\text{orig}, \tau_\text{perfect}) = 1$. The resulting forward pair is
\[
P_\text{orig}: \tau_\text{perfect} \succ \tau_\text{fail}.
\]
In our main instantiation, this process yields 1,443 forward pairs.

\subsection{Reverse Intent Anchor}

The reverse anchor is added only when the failure can safely be interpreted as a correct response to another truthful intent. We use a conservative pipeline: outcome-summary eligibility, achieved-prompt synthesis, and cycle-consistency validation. Failures that do not pass remain forward-only.

\paragraph{Outcome-summary eligibility.}
Before generating any achieved prompt, the teacher summarizes what $\tau_\text{fail}$ accomplishes and labels it \emph{eligible}, \emph{forward-only}, or \emph{reject-all}. Eligible failures must be coherent, substantially complete, and mainly attributable to instruction mismatch (e.g., language, format, length, structure, or keyword constraints). We exclude factual, mathematical, or logical errors; hallucinations; unsafe content; degeneration; incoherence; and incomplete task content.

\paragraph{Achieved prompt synthesis.}
For each eligible failure, the teacher synthesizes a standalone \emph{achieved prompt} $P_\text{achieved}$ under which $\tau_\text{fail}$ should be correct. This prompt cannot make the failure correct by merely deleting constraints; it must replace violated constraints with the content, language, format, tone, length, and structure actually exhibited by $\tau_\text{fail}$, while avoiding leakage, false content, meta-evaluation wording, or copy requests. We audit naturalness and leakage in the supplementary material.

\paragraph{Cycle-consistency validation.}
A separate teacher judge validates the completed crossed anchor under $P_\text{achieved}$. A reverse pair is accepted only if the achieved prompt is non-leaky, $\tau_\text{fail}$ is coherent, truthful, safe, and complete, $\tau_\text{perfect}$ is less appropriate under $P_\text{achieved}$, and the preference is clear. Accepted examples form the reverse pair
\[
P_\text{achieved}: \tau_\text{fail} \succ \tau_\text{perfect}.
\]
We prefer precision over coverage: 282 of 316 candidates pass validation (89.2\%). Prompt-template summaries and the validation rubric are provided in the supplementary material.

\subsection{DPO Training on Mixed BPS Dataset}

BPS does not modify the DPO objective. After constructing forward and reverse anchors, we form the mixed preference dataset
\begin{equation}
\mathcal{D}_{\text{BPS}} = \mathcal{D}_{\text{forward}} \cup \mathcal{D}_{\text{reverse}},
\end{equation}
where forward pairs have the form $(P_{\text{orig}}, \tau_{\text{perfect}}, \tau_{\text{fail}})$ and validated reverse pairs have the form $(P_{\text{achieved}}, \tau_{\text{fail}}, \tau_{\text{perfect}})$. We then optimize the standard DPO objective over this mixed dataset, using a frozen reference policy $\pi_{\text{ref}}$ (instantiated as $\pi_{\theta_0}$ in our experiments):
\begin{multline}
\mathcal{L}_{\text{BPS}}
\!=\! \mathbb{E}_{(x,y_w,y_l)\sim \mathcal{D}_{\text{BPS}}}
\!\Big[
\!-\!\log \sigma\!\Big(
\beta \!\Big[
\log\frac{\pi_\theta(y_w|x)}{\pi_{\text{ref}}(y_w|x)}
\\
- \log\frac{\pi_\theta(y_l|x)}{\pi_{\text{ref}}(y_l|x)}
\Big]\Big)
\Big].
\end{multline}
Each preference pair receives the same per-sample weight. The aggregate strength of the reverse signal is therefore controlled by the reverse density $\rho = |\mathcal{D}_{\text{reverse}}| / |\mathcal{D}_{\text{forward}}|$. Our final density, $\rho \approx 0.195$ (1,443 forward + 282 reverse = 1,725 total pairs), is determined by retaining all reverse pairs that pass validation; sensitivity checks around this naturally resulting density are reported in the supplementary material. Thus, BPS changes the preference supervision available to DPO, but not the optimization objective itself.

\section{Experiments and Analysis}

\subsection{Setup}

\paragraph{Model.}
We use \textbf{Qwen3-4B-Instruct-2507}~\cite{yang2025qwen3} as the student and a frozen copy as the DPO reference policy. \textbf{Gemini-3.1-pro-preview} is used only for offline hint generation, achieved-prompt synthesis, and validation.

\paragraph{Data.}
Seed prompts are drawn from \textbf{IFBench-train}~\cite{pyatkin2025generalizingverifiableinstructionfollowing}; evaluation prompts are excluded, and our n-gram overlap checks find no exact or near-duplicate generated $P_\text{achieved}$ prompts in the evaluated prompt sets. After dual-temperature rollouts and filtering, we obtain \textbf{1,443 forward pairs} and \textbf{282 reverse pairs} (1,725 total, $\rho \approx 0.195$). About 71\% of initial failures have $r \geq 0.5$.

\paragraph{Teacher usage and quality control.}
The teacher is not used for downstream scoring. The main construction run uses about 2,500 teacher calls / 5M tokens. Strict validation accepts 282 of 316 pre-filtered reverse candidates and discards ambiguous cases. The Gemini held-out crossed-anchor probe is therefore treated as a reward-geometry diagnostic rather than teacher-independent generalization; we add a Kimi-K2.6 synthesized-and-validated probe as a cross-teacher check.

\paragraph{Training.}
All models follow the same standard-DPO training configuration ($\beta=0.1$) and three-epoch schedule; the main comparison uses the common step-100 checkpoint, matching the number of optimizer updates. Full hyperparameters are in the supplementary material.

\paragraph{Baselines.}
\textbf{Base}: Qwen3-4B-Instruct-2507 without further training.
\textbf{Forward-DPO} (Step 60 / Step 100): Trained on $\mathcal{D}_\text{forward}$ only. Fwd-100 is the checkpoint matched to the main BPS comparison; Fwd-60 is an earlier checkpoint evaluated in the supplementary material.
\textbf{BPS} (BPS-100): Trained on $\mathcal{D}_\text{forward} \cup \mathcal{D}_\text{reverse}$, with additional checkpoints used for training-dynamics analysis (Section~\ref{sec:dynamics}).
\textbf{Shuffled-Rev} (Shuf): An ablation that keeps each reverse response pair and label direction but randomly permutes achieved prompts across examples, used only in analysis (Section~\ref{sec:ablation}).
\textbf{Constraint-Rev} (C-Rev): A strong RPO-inspired construction baseline that uses the same boundary-failure pool and reverse-pair budget as BPS, but creates reverse prompts by mechanically modifying violated constraints in the original prompt rather than synthesizing the achieved intent.

\paragraph{Benchmarks.} We evaluate instruction following (IFBench~\cite{pyatkin2025generalizingverifiableinstructionfollowing}, IFEval~\cite{zhou2023instruction}, FollowBench~\cite{jiang2024followbench}, Multi-IF~\cite{he2024multi}), reasoning and math (GPQA-Diamond~\cite{rein2023gpqa}, MATH-500~\cite{hendrycks2021measuring}), code (HumanEval+~\cite{chen2021evaluating,liu2023your}, MBPP~\cite{austin2021program}), Chinese language (C-Eval~\cite{huang2023c}), agent/tool use (tau-bench~\cite{yao2024taubenchbenchmarktoolagentuserinteraction}, tau2-bench~\cite{barres2025tau2benchevaluatingconversationalagents}, ToolBench~\cite{qin2023toolllmfacilitatinglargelanguage}), and broad reasoning (MMLU-Pro~\cite{wang2024mmlu}, BBH~\cite{suzgun2023challenging}). FollowBench reports CSL/HSR/SSR in English and Chinese; Multi-IF averages eight languages over three turns. Full metrics and evaluation details are in the supplementary material. All models use the same decoding configuration and official/default evaluators.

\noindent We focus on both downstream behavior and learned preference geometry. Section~\ref{sec:fc_boundary_main} reports a separate function-calling instantiation of the BPS pipeline; because it uses different boundary data and checkpoint selection, its numbers are not mixed with the main IFBench-trained comparisons.

\subsection{Main Results}

Table~\ref{tab:main} reports the main instruction-following results at the matched step-100 checkpoint.

\begin{table*}[t]
\centering
{
\small
\begin{tabular*}{\textwidth}{@{\extracolsep{\fill}}lccc@{\hspace{1.2em}}lccc}
\toprule
\multicolumn{4}{c}{\textbf{IFBench / IFEval}}
& \multicolumn{4}{c}{\textbf{FollowBench / Multi-IF}} \\
\cmidrule(lr){1-4}\cmidrule(lr){5-8}
\textbf{Metric} & \textbf{Base} & \textbf{Fwd-100} & \textbf{BPS-100}
& \textbf{Metric} & \textbf{Base} & \textbf{Fwd-100} & \textbf{BPS-100} \\
\midrule
IFBench-P Strict  & 31.00 & 31.97 & \textbf{32.31}
& FollowBench CSL (EN)  & \textbf{3.00} & 2.96  & \textbf{3.00} \\
IFBench-I Strict  & 33.67 & \textbf{36.05} & 35.71
& FollowBench CSL (ZH)  & 2.92          & 2.90  & \textbf{2.98} \\
IFBench-P Loose   & 33.33 & 32.65 & \textbf{35.03}
& FollowBench HSR (EN)  & 66.04         & 64.44 & \textbf{66.37} \\
IFBench-I Loose   & 36.50 & 36.90 & \textbf{38.78}
& FollowBench HSR (ZH)  & 64.46         & 64.36 & \textbf{65.47} \\
\cmidrule(lr){1-4}
IFEval-P Strict   & 82.81 & 83.36 & \textbf{86.69}
& FollowBench SSR (EN)  & 74.88         & 74.62 & \textbf{75.24} \\
IFEval-I Strict   & 88.88 & 90.36 & \textbf{91.87}
& FollowBench SSR (ZH)  & 72.36         & 72.99 & \textbf{74.58} \\
IFEval-P Loose    & 86.14 & 84.66 & \textbf{89.28}
& Multi-IF Turn 1 Avg.  & 85.72         & 87.21 & \textbf{87.93} \\
IFEval-I Loose    & 90.88 & 91.04 & \textbf{93.25}
& Multi-IF Turn 2 Avg.  & 78.07         & 75.78 & \textbf{79.15} \\
\cmidrule(lr){1-4}
IFBench Avg.      & 33.63 & 34.39 & \textbf{35.46}
& Multi-IF Turn 3 Avg.  & 69.45         & 66.14 & \textbf{70.48} \\
IFEval Avg.       & 87.18 & 87.36 & \textbf{90.27}
& Multi-IF Overall Avg. & 77.75         & 76.38 & \textbf{79.19} \\
\bottomrule
\end{tabular*}
}
\caption{Instruction-following results at Step 100. IFBench/IFEval averages are means over listed metrics; Multi-IF scores are macro-averaged across eight languages. BPS-100 scores above both comparisons at every Multi-IF turn, while the BPS-100--Fwd-100 gap grows monotonically with turn depth (+0.72, +3.37, +4.34); per-language analysis is in the supplementary material.}
\label{tab:main}
\end{table*}

Table~\ref{tab:main} shows the main capability-retention trade-off. BPS scores above Base on all listed IFBench and IFEval variants, attains the strongest listed FollowBench scores, and scores above both comparison models at every Multi-IF turn. Forward-DPO improves some shallow instruction-following metrics but is below Base on most FollowBench metrics and later Multi-IF turns. A separate three-seed check also shows a consistent IFEval gain for BPS (87.79$\pm$0.23 for Forward-DPO vs.\ 88.83$\pm$0.59 for BPS; see the supplementary material).
We compare BPS and Forward-DPO after the same number of optimizer updates at step 100. An earlier Forward-DPO checkpoint exhibits a different narrow-performance/capability-retention trade-off and is reported as a checkpoint-sensitivity comparison in the supplementary material.

\paragraph{Multi-turn multilingual robustness.}
Multi-IF makes the depth effect clear. Relative to Base, Fwd-100 changes from $+$1.49 on turn 1 to $-$2.29 and $-$3.31 on turns 2--3, while BPS improves all turns ($+$2.21, $+$1.08, $+$1.03). The BPS--Fwd gap therefore grows with turn depth ($+$0.72, $+$3.37, $+$4.34), and the per-language breakdown in the supplementary material shows the same pattern in 23/24 language-turn averages.

\paragraph{General and out-of-distribution checks.}
\label{sec:ood}

Table~\ref{tab:capability_ood} reports selected general-capability, agentic, tool-use, and broad-reasoning checks that probe whether preference training preserves capabilities outside the seed instruction-following distribution. Complete benchmark metrics are reported in the supplementary material.

\begin{table*}[t]
\centering
{
\small
\begin{tabular*}{\textwidth}{@{\extracolsep{\fill}}lccc@{\hspace{1em}}lccc}
\toprule
\multicolumn{4}{c}{\textbf{General Capabilities}} &
\multicolumn{4}{c}{\textbf{Agent, Tool Use, and Broad Reasoning}} \\
\cmidrule(lr){1-4}\cmidrule(lr){5-8}
\textbf{Metric} & \textbf{Base} & \textbf{Fwd} & \textbf{BPS}
& \textbf{Metric} & \textbf{Base} & \textbf{Fwd} & \textbf{BPS} \\
\midrule
GPQA-Diamond      & 42.93 & 45.96 & \textbf{46.46}
& tau-bench airline      & \textbf{36.00} & 30.00 & \textbf{36.00} \\
MATH-500          & 94.80 & 94.00 & \textbf{95.00}
& tau2-bench airline     & \textbf{27.91} & 25.58 & \textbf{27.91} \\
HumanEval+        & 71.34 & 71.34 & \textbf{72.56}
& tau2-bench retail      & \textbf{43.75} & 36.61 & \textbf{43.75} \\
MBPP              & 72.00 & 72.00 & \textbf{72.80}
& ToolBench Act.EM (ov.) & \textbf{35.50} & 33.10 & 35.46 \\
C-Eval Overall    & \textbf{80.39} & 78.75 & \textbf{80.39}
& ToolBench Plan EM (OOD)& 78.50 & 77.98 & \textbf{81.06} \\
BBH Overall       & 86.74 & 84.27 & \textbf{86.96}
& ToolBench HalluRate (OOD)$\downarrow$ & \textbf{0.53} & 0.66 & \textbf{0.53} \\
\bottomrule
\end{tabular*}
}
\caption{Selected capability-retention and out-of-distribution checks at Step 100. BPS scores at or above Base on the listed general-capability metrics, recovers Base-level scores on the shown agentic tasks where Fwd-100 is lower, and matches the Base hallucination rate on ToolBench OOD. Complete ToolBench and additional benchmark metrics are in the supplementary material.}
\label{tab:capability_ood}
\end{table*}

In this run, BPS is above Base on the listed GPQA-Diamond~\cite{rein2023gpqa}, MATH-500~\cite{hendrycks2021measuring}, HumanEval+, MBPP, and BBH scores and matches Base on C-Eval. On the shown agentic and tool-use settings, Forward-DPO is below Base on several metrics and has a higher ToolBench OOD hallucination rate; BPS recovers these metrics toward Base, nearly matches Base ToolBench Act.EM, matches its hallucination rate, and attains the strongest ToolBench OOD Plan EM. Together with the instruction-following results, these checks support the practical goal of improving correction behavior while retaining useful capabilities for multilingual interaction, tool use, coding, and general reasoning.

\paragraph{Additional model checks.}
Additional Qwen-family checks using IFEval prompts as seeds show the same IFBench pattern and improved code metrics in those settings, while leaving MMLU stable (see the supplementary material).

\paragraph{Function-calling pipeline instantiation.}
\label{sec:fc_boundary_main}

We further test whether BPS transfers beyond surface instruction constraints by rebuilding the complete pipeline in a function-calling setting, where an argument value can be wrong for one request but correct for a nearby request. This run uses separate xLAM-style single-call data~\cite{zhang2025xlam}, achieved prompts, validation filters, and DPO pairs; it is therefore a domain-transfer test of the data-construction method rather than an evaluation of the IFBench-trained checkpoint.

Candidates are parseable, schema-valid calls to the gold function that differ only in argument values. A teacher validates both directions:
\[
P_\text{orig}: y_\text{gold} \succ y_\text{model}, \quad
P_\text{achieved}: y_\text{model} \succ y_\text{gold}.
\]
The validated pool contains 342 forward examples and 231 reverse candidates; we train with $\rho=0.15$ reverse density and use the final checkpoint (step 60) for both function-calling Forward-DPO and BPS.

\begin{table}[t]
\centering
{
\small
\begin{tabular}{lcc}
\toprule
\textbf{Metric} & \textbf{Fwd-DPO} & \textbf{BPS} \\
\midrule
FC Full Acc. & 0.860 & \textbf{0.869} \\
Acc$_{\text{achieved}}$ & 0.759 & \textbf{0.824} \\
Boundary Ach EM & 0.721 & \textbf{0.768} \\
Boundary Both EM & 0.443 & \textbf{0.489} \\
\bottomrule
\end{tabular}
}
\caption{Function-calling pipeline instantiation (ckpt-60). BPS has a consistent directional advantage on held-out full-call accuracy and achieved-side boundary switching when the full pipeline is rebuilt for structured function calls.}
\label{tab:function_calling_probe}
\end{table}

Table~\ref{tab:function_calling_probe} shows a consistent directional advantage for BPS across full-call accuracy, achieved-side preference ranking, achieved-prompt exact match, and joint boundary success. This separate instantiation provides evidence that crossed-anchor construction can be applied to structured action spaces, not only natural-language formatting constraints. Full function-calling, agentic, and code results for this separate instantiation are in the supplementary material.

\subsection{Diagnostic Analysis}
\label{sec:dynamics}
\label{sec:conditional_reward}

\paragraph{Training dynamics.}
\begin{figure}[t]
\centering
\includegraphics[width=0.95\columnwidth]{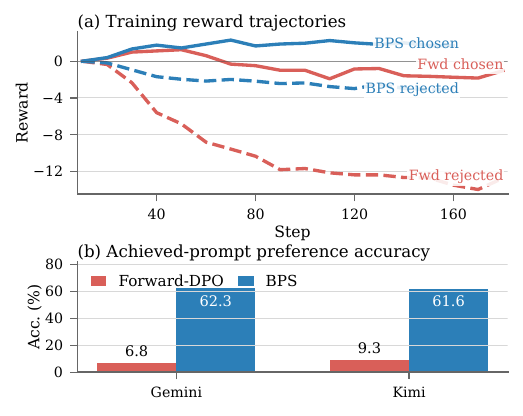}
\caption{Diagnostic view of BPS. (a) Forward-DPO expands DPO margins mainly by lowering rejected rewards, while BPS keeps chosen rewards positive and rejected suppression more moderate. (b) BPS substantially raises achieved-side pairwise ranking accuracy on both held-out probes.}
\label{fig:dynamics}
\end{figure}

Figure~\ref{fig:dynamics} shows that Forward-DPO reduces loss largely by driving rejected rewards downward: by step 60 the rejected reward is $-8.83$, and by step 100 the chosen reward is also negative. BPS keeps chosen rewards positive and rejected suppression more moderate. Because aggregate dynamics can reflect data composition, length, and optimization differences, we treat them only as diagnostic and next test held-out crossed anchors directly.

\paragraph{Conditional reward geometry.}

\begin{table*}[t]
\centering
{
\small
\begin{tabular}{llcccc}
\toprule
\textbf{Probe} & \textbf{Model} & $M_o$ & Acc$_o$ & $M_a$ & Acc$_a$ \\
\midrule
\multirow{2}{*}{Gemini}
& Forward-DPO & 0.540 & 99.3 & $-$0.349 & 6.8 \\
& BPS         & 0.292 & 100.0 & \phantom{$-$}0.011 & 62.3 \\
\midrule
\multirow{2}{*}{Kimi}
& Forward-DPO & 0.854 & 99.3 & $-$0.374 & 9.3 \\
& BPS         & 0.663 & 99.3 & $-$0.005 & 61.6 \\
\bottomrule
\end{tabular}
}
\caption{Conditional reward geometry on held-out crossed anchors. Gemini denotes the original pipeline-held-out probe with 146 anchors; Kimi denotes a cross-teacher probe with 151 anchors synthesized and validated by Kimi-K2.6. BPS preserves original-side pairwise ranking while substantially improving achieved-side pairwise ranking accuracy under both construction pipelines. On the Kimi probe, the paired achieved-margin improvement over Forward-DPO is $+0.369$ (95\% bootstrap CI $[0.267,0.487]$).}
\label{tab:conditional_reward}
\end{table*}

On 146 Gemini pipeline-held-out crossed anchors, we report token-normalized DPO implicit-reward margins $M_\text{orig}$ ($\tau_\text{perfect}$ over $\tau_\text{fail}$ under $P_\text{orig}$) and $M_\text{achieved}$ (the reverse ordering under $P_\text{achieved}$), plus the fraction with the intended sign. Here accuracy denotes pairwise ranking, not generation accuracy.

Table~\ref{tab:conditional_reward} shows that both models preserve the original-side ranking. The difference is under $P_\text{achieved}$: Forward-DPO assigns the intended achieved-side ordering on 6.8\% of Gemini anchors, whereas BPS raises this pairwise ranking accuracy to 62.3\%, with a paired achieved-margin gain of $+0.360$ (95\% CI $[0.293,0.435]$). The absolute BPS achieved margin remains small, so we treat this as a clear shift in reward geometry rather than claiming complete preference reversal.
This pattern is stable across three training seeds: on the Gemini crossed-anchor probe, BPS raises achieved-side pairwise ranking accuracy from $8.0\pm0.6$\% to $65.5\pm2.6$\%, with pooled $\Delta M_\text{achieved}=+0.360$ (95\% CI $[0.320,0.401]$; see the supplementary material).

\paragraph{Cross-teacher Kimi probe.}
The Kimi-K2.6 cross-teacher probe shows the same pattern: BPS raises achieved-side pairwise ranking accuracy from 9.3\% to 61.6\%, with paired achieved-margin improvement $+0.369$ (95\% CI $[0.267,0.487]$). This provides a construction-model robustness check complementary to the blind human audit below.

\paragraph{Human audit of achieved prompts.}
Because automatic LLM judgments can be biased~\cite{zheng2023judging,liu2023g}, we conduct a blind human audit of 200 held-out anchors (100 Gemini-source and 100 Kimi-source; three annotators per anchor) to check whether labels are only a teacher self-consistency artifact. Annotators see only $P_\text{achieved}$ and two shuffled responses. Majority vote finds 91.0\% natural achieved prompts, 99.0\% agreement with the reverse preference direction, and 0.5\% leakage. Full audit results are in the supplementary material.

\subsection{Ablation Studies}
\label{sec:ablation}

\paragraph{Reverse-data ablations.}
A possible alternative is that reverse pairs merely regularize DPO. We test this with Shuffled-Rev, which preserves reverse-pair count and labels but destroys semantic prompt-response alignment:
\[
(P_\text{achieved}^{\pi(i)},\ \tau_\text{fail}^i \succ \tau_\text{perfect}^i), \quad \pi(i) \neq i.
\]
Constraint-Rev is a strong RPO-inspired baseline that uses the same boundary-failure pool and reverse-pair budget as BPS, but mechanically modifies violated constraints in $P_\text{orig}$ without summarizing the intent actually achieved by the failure.

\begin{table}[t]
\centering
{
\footnotesize
\begin{tabular}{lccccc}
\toprule
\textbf{Metric} & \textbf{Base} & \textbf{Fwd} & \textbf{Shuf} & \textbf{C-Rev} & \textbf{BPS} \\
\midrule
IFBench Avg.    & 33.63 & 34.39 & 33.16 & 32.61 & \textbf{35.46} \\
IFEval Avg.     & 87.18 & 87.36 & 88.28 & 88.62 & \textbf{90.27} \\
M-IF Turn 1     & 85.72 & 87.21 & 87.57 & 86.89 & \textbf{87.93} \\
M-IF Turn 2     & 78.07 & 75.78 & 77.44 & 77.94 & \textbf{79.15} \\
M-IF Turn 3     & 69.45 & 66.14 & 68.32 & 69.26 & \textbf{70.48} \\
\midrule
Reward Margin   & --    & 10.71 & 6.71 & 2.81 & 4.33  \\
\bottomrule
\end{tabular}
}
\caption{Reverse-data ablations at step 100. Shuf permutes achieved prompts across reverse pairs. C-Rev is a strong RPO-style construction baseline that mechanically reverses violated constraints without achieved-intent synthesis. Both reverse-style baselines mitigate Forward-DPO's multi-turn degradation, while BPS gives the strongest balance across IFBench, IFEval, and Multi-IF in this comparison.}
\label{tab:shuffle_ablation}
\end{table}

Table~\ref{tab:shuffle_ablation} shows that shuffled reverse partly reduces Forward-DPO's later-turn degradation but weakens IFBench, while Constraint-Rev helps later Multi-IF turns but drops IFBench below Base. Among the tested constructions, BPS gives the strongest observed balance across IFBench, IFEval, and Multi-IF.

\paragraph{Budget-matched reward controls.}
Budget-matched controls further isolate the role of crossed achieved-intent anchors (see the supplementary material). Duplicating 282 forward pairs leaves achieved-side ranking accuracy unchanged at 6.8\%, ruling out extra optimizer exposure as an explanation for the observed shift. Adding 282 mechanically reversed constraint pairs raises achieved-side ranking accuracy to 32.9\%, but remains well below BPS at 62.3\% and weakens original-side accuracy (95.9\% vs.\ 100.0\%). Thus, reverse supervision helps only partially without achieved-intent matching; neither extra pair count nor reverse labels alone reproduces the BPS result.

The Kimi-heldout reward-geometry ablation agrees: Constraint-Rev raises achieved-side ranking accuracy over Forward-DPO but remains below BPS, and BPS gives a positive paired achieved-margin gain over Constraint-Rev ($+0.066$, 95\% CI $[0.043,0.095]$; see the supplementary material).

\paragraph{Interpreting the crossed-anchor effect.}

Forward pairs lower $\tau_\text{fail}$ relative to $\tau_\text{perfect}$ under $P_\text{orig}$, and shared parameters allow this update to affect other prompts. Forward-DPO's low achieved-side rankings are consistent with such cross-prompt rejection. BPS supplies the complementary signal $P_\text{achieved}: \tau_\text{fail} \succ \tau_\text{perfect}$, retaining the original-side ordering while learning a different achieved-side ordering. Density sensitivity shows the expected trade-off: $\rho=0.10$ gives a smaller correction, whereas $\rho=0.30$ reduces IFBench; the validation-determined $\rho\approx0.195$ lies between them (supplementary material).

This interpretation aligns with the ablations: reverse labels work best when the prompt captures the response mode expressed by the failure. BPS therefore records correction and preservation as prompt-dependent facts about the same response pair.

\paragraph{No label contradiction.} Preferences are over prompt-response pairs: $\tau_\text{perfect}\succ\tau_\text{fail}$ under $P_\text{orig}$, while $\tau_\text{fail}\succ\tau_\text{perfect}$ under $P_\text{achieved}$. BPS therefore teaches a prompt-conditioned boundary, not contradictory labels.

\paragraph{Boundary failure selection.}

Near misses ($r \geq 0.5$) are suitable reverse candidates because they often remain coherent despite localized constraint mismatches; 71\% of our boundary failures satisfy this criterion. Most validated anchors involve format/structure or lexical mismatches, and BPS raises achieved-side ranking accuracy across all observed types (supplementary material).

\paragraph{Verbosity.}
Forward-DPO is also associated with substantially longer outputs: IFBench average length rises from 1,584 tokens (Base) to 5,920, while BPS is 2,565; on ToolBench, BPS stays closer to Base. This pattern is consistent with length being used as a constraint-coverage strategy~\cite{dubois2024length}; full statistics are in the supplementary material.

\section{Conclusion}
Forward-only preference data can treat coherent boundary failures as intrinsically negative. BPS instead adds validated achieved-prompt reverse anchors, encoding where the same response is wrong and where it is right without changing the DPO objective.

Across held-out reward probes and downstream evaluations, BPS counteracts cross-prompt rejection patterns while preserving the original preference boundary. Its clearest downstream separation appears in multilingual multi-turn instruction following, with consistent capability-retention evidence from tool-use, agentic, and code checks. The results support encoding not only which responses are wrong, but also the prompts under which they are useful.

\appendix
\section*{Supplementary Material}

\section{Limitations}

\paragraph{Achieved-intent representation.}
BPS represents the achieved intent as a natural-language prompt so that the resulting data can be used by standard preference-optimization pipelines without changing the model architecture or DPO objective. In domains with formal action spaces, such as function calling or tool use, implementations may also store schemas or constraint lists alongside the achieved prompt for easier inspection.

\paragraph{Offline construction.}
BPS shifts effort from online reinforcement learning to offline data construction. Our implementation uses teacher calls for correction hints, achieved-prompt synthesis, and validation, plus an optional human audit. These steps are independent of DPO training and can be batched, cached, or replaced with cheaper verifier models in larger runs.

\paragraph{Domain adaptation.}
The same BPS recipe applies to both instruction following and function calling in our experiments, but each domain benefits from validation criteria that match its output structure. Instruction-following data uses constraint satisfaction and leakage checks, while function-calling data also checks parseability, schema validity, function identity, and argument values.

\paragraph{Objective scope.}
We study BPS with standard DPO to isolate the effect of crossed boundary-failure pairs. The same data construction could be combined with objective-level improvements such as DPO-positive/Smaug, RPO/IOPO-style variants, or other preference-optimization losses.

\section{Ethical Statement}

This work studies an offline data-construction method for preference optimization. We do not use private user conversations or data from deployed systems. The only newly collected human judgments are task-focused annotations used for the achieved-prompt audit; the annotation task contains no personal or sensitive user content. The seed prompts, evaluation benchmarks, and model checkpoints used in this study are public research artifacts, and we use them only for research purposes under their respective licenses and terms of use. We do not redistribute third-party benchmark test sets or model weights.

BPS constructs additional preference pairs by synthesizing achieved prompts for boundary failures. Our pipeline keeps reverse pairs only after eligibility, safety, truthfulness, and non-leakage validation, and discards ambiguous cases. These filters are intended to prevent private, unsafe, hateful, deceptive, hallucinated, or factually incorrect outputs from being preserved as positives.

Achieved-prompt synthesis uses an offline teacher model, so applications should document the teacher, validation rubric, and data provenance. For multilingual, culturally sensitive, or safety-relevant data, additional audit can be applied using the same blind-review protocol used in our experiments.

Our experiments involve offline teacher-model inference and GPU fine-tuning. We report teacher-call and compute estimates to support reproducibility and to make the construction cost transparent.

\section{AI Assistance Statement}
We used ChatGPT for language polishing only. The scientific claims, experimental design, data construction pipeline, analysis, and reported results were produced and verified by the authors. As described in Sections~3 and~4, Gemini-3.1-pro-preview was used as an offline teacher for correction hints, achieved-prompt synthesis, and validation; it was not used for downstream benchmark scoring.

\section{Artifacts, Licenses, and Intended Use}
\label{sec:artifacts_licenses}

We use several third-party artifacts, including Qwen model checkpoints~\cite{yang2025qwen3,qwen2025qwen25technicalreport}, IFBench-train seed prompts~\cite{pyatkin2025generalizingverifiableinstructionfollowing}, public evaluation benchmarks, xLAM-style function-calling data~\cite{zhang2025xlam}, official benchmark evaluators, and the DeepSpeed training framework~\cite{rasley2020deepspeed}. We cite the creators of these artifacts in the main text or supplementary material.

We reviewed the licenses and terms of use from the official model cards, dataset repositories, benchmark websites, or associated papers. All artifacts are used for research purposes only. We do not redistribute original benchmark test sets or third-party model weights. Any released code, metadata, or derived BPS data will be distributed only when permitted by the corresponding upstream license or terms, and users will be instructed to comply with the original artifact licenses.

Our use is consistent with the intended research use of the artifacts. IFBench-train is used only as a seed source for research data construction, public benchmarks are used only for evaluation under their official protocols, and Qwen checkpoints are used as research baselines and fine-tuning targets under their released terms. BPS-derived artifacts are intended for research on preference learning and instruction-following alignment, not for safety-critical or high-stakes deployment without further validation.

Because BPS can derive new preference pairs from existing prompts and model outputs, any redistribution of such derived data should preserve the access conditions of the source artifacts. If an upstream artifact permits research use but restricts redistribution or commercial use, the corresponding derived examples should be used only under compatible conditions.

\section{Full Results for Function-Calling Agentic Generalization}
\label{sec:fc_agent_full}

Table~\ref{tab:fc_agent_full} reports the complete per-split results for the function-calling instantiation described in the main paper. This instantiation rebuilds the full BPS data-construction pipeline in the function-calling domain using Gemini as the teacher, two-stage validation, 342 forward pairs, and a reverse density of $\rho=0.15$ (51 reverse pairs).

\begin{table*}[t]
\centering
{
\small
\begin{tabular}{llccc}
\toprule
\textbf{Benchmark} & \textbf{Metric / Split} & \textbf{Base} & \textbf{FC-Fwd} & \textbf{FC-BPS} \\
\midrule
General-FC & Tool-call F1      & 52.9 & 53.7 & \textbf{54.2} \\
General-FC & Schema Acc.       & 67.8 & 68.2 & \textbf{68.8} \\
General-FC & Successful Calls  & 396  & 401  & \textbf{411} \\
\midrule
$\tau$-bench & Airline Acc.    & 36.0 & 26.0 & \textbf{42.0} \\
$\tau$-bench & Retail Acc.     & \textbf{55.7} & 46.1 & 50.4 \\
$\tau$-bench & Overall Acc.    & \textbf{49.7} & 40.0 & 47.9 \\
\midrule
$\tau^2$-bench & Airline Acc.  & 27.9 & 27.9 & \textbf{37.2} \\
$\tau^2$-bench & Retail Acc.   & 32.1 & \textbf{43.8} & 39.3 \\
$\tau^2$-bench & Telecom Acc.  & 35.1 & 30.7 & \textbf{35.1} \\
$\tau^2$-bench & Overall Acc.  & 34.2 & 35.7 & \textbf{37.2} \\
\midrule
HumanEval+ & Pass@1            & 71.3 & \textbf{72.6} & \textbf{72.6} \\
MBPP+ & Pass@1                 & 82.0 & 82.0 & \textbf{82.3} \\
MBPP  & Pass@1                 & 72.0 & 70.6 & \textbf{72.4} \\
\bottomrule
\end{tabular}
}
\caption{Full results for the separate function-calling instantiation (Base = Qwen3-4B without further training). Scores are percentages except successful-call counts. BPS improves function-calling accuracy, $\tau^2$-bench overall, and MBPP+ over both Base and FC-Fwd. On $\tau$-bench, BPS outperforms FC-Fwd on airline and overall but falls below Base on retail, yielding a slightly lower overall score than Base. The main exception across benchmarks is $\tau^2$-retail, where FC-Fwd scores higher.}
\label{tab:fc_agent_full}
\end{table*}

\section{Training Details}
\label{sec:training_details}

All models are trained with standard DPO (sigmoid loss, $\beta=0.1$) using full fine-tuning via DeepSpeed ZeRO-2~\cite{rasley2020deepspeed}, learning rate $1\times10^{-6}$ with cosine schedule, per-device batch size 4, global batch size 32, \texttt{bf16} precision, and a maximum sequence length of 1,536 tokens. The sequence length limit applies only to DPO training; evaluation uses each benchmark's decoding budget, allowing longer generations. Training runs for 3 epochs ($\approx$160 optimizer steps on the BPS-sized dataset). Main results use a fixed step-100 comparison rather than validation-selected best checkpoints; Fwd-60 is reported separately for checkpoint-sensitivity analysis.

\paragraph{Compute and teacher calls.}
All DPO experiments were run on 8$\times$ NVIDIA A800 GPUs with 80GB memory each. A main DPO run took approximately 0.2 hours, corresponding to 1.6 GPU-hours. Including Forward-DPO, BPS, reverse-density ablations, shuffled-reverse ablation, and function-calling runs, the total fine-tuning budget was approximately 15 GPU-hours. Teacher-model inference for data construction required approximately 2,500 calls / 5M tokens and is not included in the GPU-hour estimate. Thus, BPS is lightweight on the training side but shifts cost to offline data construction; reducing teacher cost is an implementation question for larger-scale use.

\section{Evaluation Implementation Details}
\label{sec:evaluation_details}

We use the official evaluation scripts or released benchmark evaluators whenever available. All models are evaluated with the same decoding configuration: temperature 0.7, top-p 0.9, and maximum generation length 8192 tokens. For benchmarks with official default settings, we follow the default aggregate metric and reporting protocol. For IFBench, IFEval, FollowBench, Multi-IF, ToolBench, tau-bench, tau2-bench, HumanEval+, MBPP+, GPQA-Diamond~\cite{rein2023gpqa}, MATH-500~\cite{hendrycks2021measuring}, C-Eval, MMLU-Pro, and BBH, we use the corresponding official or commonly adopted evaluation implementations. Exact-match, pass@1, F1, hallucination-rate, and instruction-satisfaction metrics are computed according to each benchmark's official definition.

\section{Data Construction Funnel}
\label{sec:data_funnel}

Table~\ref{tab:data_funnel} summarizes the retention rates at each stage of the BPS data construction pipeline.

\begin{table}[t]
\centering
{
\small
\begin{tabular}{lrr}
\toprule
\textbf{Stage} & \textbf{Count} & \textbf{Retention} \\
\midrule
Seed prompts (IFBench-train)        & 2{,}000 & --- \\
Student rollouts (dual-temperature) & 6{,}000 & --- \\
Boundary failures ($0 < r < 1$)     & 3{,}420 & 57.0\% \\
Near-miss failures ($r \geq 0.5$)   & 2{,}428 & 71.0\% \\
Accepted forward pairs              & 1{,}443 & 59.4\% \\
Reverse candidates (pre-filtered)   &   316   & 13.0\% \\
Valid reverse pairs                 &   282   & 89.2\% \\
\midrule
\textbf{Total DPO pairs}            & \textbf{1{,}725} & --- \\
\bottomrule
\end{tabular}
}
\caption{Data construction funnel. Retention rates are relative to the immediately preceding stage, except ``Reverse candidates'' (relative to near-miss failures) and ``Valid reverse pairs'' (relative to reverse candidates). Dual-temperature decoding samples $K=3$ rollouts per prompt.}
\label{tab:data_funnel}
\end{table}

\section{Full ToolBench Results}
\label{sec:toolbench_full}

Table~\ref{tab:toolbench_full} presents the full ToolBench results across all splits and metrics.

\begin{table*}[t]
\centering
{
\footnotesize
\begin{tabular}{lccc}
\toprule
\textbf{Benchmark} & \textbf{Base} & \textbf{Fwd-100} & \textbf{BPS-100} \\
\midrule
ToolBench Act.EM (overall) & \textbf{35.50} & 33.10 & 35.46 \\
ToolBench Act.EM (in)      & 35.89 & 34.06 & \textbf{36.04} \\
ToolBench Act.EM (OOD)     & \textbf{34.70} & 31.17 & 34.30 \\
ToolBench F1 (in)          & 30.86 & 29.53 & \textbf{30.97} \\
ToolBench F1 (OOD)         & \textbf{29.38} & 26.75 & 28.98 \\
ToolBench Plan EM (in)     & 80.94 & 81.39 & \textbf{83.22} \\
ToolBench Plan EM (OOD)    & 78.50 & 77.98 & \textbf{81.06} \\
ToolBench HalluRate (OOD)$\downarrow$ & \textbf{0.53} & 0.66 & \textbf{0.53} \\
\bottomrule
\end{tabular}
}
\caption{Full ToolBench results (Step 100). BPS recovers nearly all ToolBench Act.EM and F1 scores, matches Base hallucination rate, and achieves the highest Plan EM on both splits.}
\label{tab:toolbench_full}
\end{table*}

\section{Reverse Density Ablation}
\label{sec:rho_ablation}

Table~\ref{tab:rho_ablation} shows how the reverse anchor density $\rho$ affects downstream performance.

\begin{table}[t]
\centering
{
\footnotesize
\begin{tabular}{lcccc}
\toprule
\textbf{$\rho$} & \textbf{IFB Avg} & \textbf{M-IF Ov.} & \textbf{LCB} & \textbf{TB OOD} \\
\midrule
0 (Fwd-100)      & 34.39 & 76.38 & 45.40 & 26.75 \\
0.10             & 35.33 & 78.61 & 46.54 & 28.41 \\
\textbf{0.195 (BPS)} & \textbf{35.46} & \textbf{79.19} & \textbf{48.53} & 28.98 \\
0.30             & 32.82 & 78.44 & 47.30 & \textbf{29.75} \\
\bottomrule
\end{tabular}
}
\caption{Reverse-density sensitivity at Step 100. IFBench Avg = mean of four metrics; M-IF = macro-avg across 8 languages/3 turns; LCB = LiveCodeBench; TB OOD = ToolBench OOD. The main density $\rho=0.195$ results from retaining all validated reverse pairs. Lower density provides a smaller correction, while $\rho=0.30$ reduces IFBench below Fwd-DPO (32.82 vs.\ 34.39), illustrating the correction--retention trade-off.}
\label{tab:rho_ablation}
\end{table}

\section{Budget-Matched Reward Controls}
\label{sec:budget_reward_controls}

Table~\ref{tab:budget_reward_controls} isolates the role of crossed achieved-intent anchors on the Gemini held-out probe. Fwd-Dup adds 282 duplicated forward pairs, matching BPS in total pair count. RPO-style adds 282 mechanically reversed constraint pairs, matching BPS in reverse-pair count but not using achieved-intent synthesis.

\begin{table}[t]
\centering
{
\footnotesize
\begin{tabular}{lcccc}
\toprule
\textbf{Model} & $M_o$ & Acc$_o$ & $M_a$ & Acc$_a$ \\
\midrule
Forward-DPO & 0.541 & 99.3 & $-$0.349 & 6.8 \\
Fwd-Dup (+282 fwd) & 0.517 & 99.3 & $-$0.334 & 6.8 \\
RPO-style (+282 rev) & 0.213 & 95.9 & $-$0.057 & 32.9 \\
BPS (+282 rev) & 0.293 & 100.0 & \phantom{$-$}0.011 & 62.3 \\
\bottomrule
\end{tabular}
}
\caption{Budget-matched reward-geometry controls on the Gemini held-out crossed-anchor probe. Duplicating forward pairs leaves achieved-side ranking accuracy unchanged. RPO-style mechanical reverse pairs partially raise achieved-side ranking accuracy but remain below BPS and weaken original-side ranking accuracy.}
\label{tab:budget_reward_controls}
\end{table}

\section{Constraint-Rev Kimi Reward Geometry}
\label{sec:constraint_rev_kimi}

Table~\ref{tab:constraint_rev_geometry} gives the Kimi-heldout conditional reward geometry for the Constraint-Rev comparison discussed in the main text.

\begin{table}[t]
\centering
{
\small
\begin{tabular}{lcccc}
\toprule
\textbf{Model} & $M_\text{orig}$ & Acc$_\text{orig}$ & $M_\text{achieved}$ & Acc$_\text{achieved}$ \\
\midrule
Forward-DPO    & 0.854 & 99.3 & $-$0.374 & 9.9 \\
Constraint-Rev & 0.485 & 97.4 & $-$0.072 & 37.1 \\
BPS            & 0.667 & 99.3 & $-$0.006 & 59.6 \\
\bottomrule
\end{tabular}
}
\caption{Kimi-heldout conditional reward geometry for Constraint-Rev, evaluated on the subset used for the Constraint-Rev comparison. Constraint-Rev raises achieved-side ranking accuracy over Forward-DPO but remains below BPS; BPS also preserves stronger original-side margins. Because this subset differs slightly from the main-paper conditional reward table, the Forward-DPO and BPS numbers differ slightly. Paired bootstrap gives $\Delta M_\text{achieved}(\text{BPS}-\text{C-Rev})=+0.066$ (95\% CI $[0.043,0.095]$).}
\label{tab:constraint_rev_geometry}
\end{table}

\section{A Challenging BPS Example}
\label{sec:bps_example_full}

Table~\ref{tab:bps_example_full} provides a translated and lightly abbreviated content-intent boundary example. We select this relatively challenging case to illustrate that the construction can handle a coherent semantic deviation in addition to the format/structure and lexical mismatches that constitute most validated anchors (Table~\ref{tab:failure_type_distribution}).

\begin{table*}[!t]
\centering
{
\small
\begin{tabular}{p{0.16\textwidth}p{0.78\textwidth}}
\toprule
\textbf{Field} & \textbf{Text} \\
\midrule
Original prompt $P_\text{orig}$
& \texttt{Which country contains the largest clock in the world? At the end of your response, please explicitly add a postscript starting with Postscript: The first word of your response should be the word grandfather. Your ENTIRE response should be in Ukrainian language, no other language is allowed. Include keyword emotion in your response.} \\
\midrule
Forward chosen $\tau_\text{perfect}$
& \textit{``grandfather [Ukrainian: The world's largest clock is in Mecca, Saudi Arabia. This clock reflects the grandeur and traditions, evoking a strong emotion in people who see it\ldots] Postscript: emotion''} \\
\midrule
Forward rejected $\tau_\text{fail}$
& \textit{``Grandfather, [Ukrainian: The largest country in the world by area is Russia. It has about 17.1 million square kilometers\ldots This evokes a strong emotion\ldots] Postscript: emotion''} \\
\midrule
Achieved prompt $P_\text{achieved}$
& \texttt{Please tell me what the largest country in the world is by area and its approximate size in square kilometers. Your response must start exactly with the English greeting ``Grandfather,''. Write the main explanation in Ukrainian. Explain how such a vast territory reflects history, culture, and nature, and mention that it evokes a strong emotion. End your response with the exact English phrase ``Postscript: emotion''.} \\
\midrule
Reverse chosen & Same as forward rejected ($\tau_\text{fail}$). \\
Reverse rejected & Same as forward chosen ($\tau_\text{perfect}$). \\
\bottomrule
\end{tabular}
}
\caption{A challenging content-intent BPS example from an IFBench-style prompt. The original prompt asks about the world's largest clock with strict multilingual and formatting constraints. The student instead gives a coherent answer about the largest country by area. The achieved prompt makes this semantic distinction explicit and is retained only after the same standalone-prompt, response-quality, and reverse-preference validation used throughout the pipeline.}
\label{tab:bps_example_full}
\end{table*}

\section{Multi-IF Full Results}
\label{sec:multiif_full}

Table~\ref{tab:multiif_full} reports the full turn-level Multi-IF metrics with the early Fwd-60 checkpoint.

\begin{table*}[!t]
\centering
{
\small
\begin{tabular}{llcccccc}
\toprule
\textbf{Turn} & \textbf{Metric} & \textbf{Base} & \textbf{Fwd-60} & \textbf{Fwd-100} & \textbf{BPS-100} & $\Delta_{\text{BPS-F60}}$ & $\Delta_{\text{BPS-F100}}$ \\
\midrule
\multirow{4}{*}{1}
& Prompt Strict & 81.81 & 83.85 & 84.47 & \textbf{84.71} & +0.86 & +0.24 \\
& Inst Strict   & 86.30 & 88.66 & \textbf{88.72} & 88.66 & +0.00 & $-$0.05 \\
& Prompt Loose  & 85.63 & 85.47 & 85.90 & \textbf{87.64} & +2.17 & +1.74 \\
& Inst Loose    & 89.13 & 89.76 & 89.76 & \textbf{90.70} & +0.94 & +0.94 \\
\midrule
\multirow{4}{*}{2}
& Prompt Strict & 67.00 & 65.45 & 65.38 & \textbf{68.87} & +3.42 & +3.49 \\
& Inst Strict   & 84.42 & 83.52 & 83.58 & \textbf{85.45} & +1.93 & +1.87 \\
& Prompt Loose  & 73.35 & 69.61 & 68.85 & \textbf{74.29} & +4.68 & +5.43 \\
& Inst Loose    & 87.52 & 85.66 & 85.33 & \textbf{88.01} & +2.35 & +2.68 \\
\midrule
\multirow{4}{*}{3}
& Prompt Strict & 51.31 & 49.01 & 48.40 & \textbf{53.20} & +4.19 & +4.80 \\
& Inst Strict   & 81.80 & 80.83 & 80.79 & \textbf{82.93} & +2.10 & +2.14 \\
& Prompt Loose  & 59.27 & 53.88 & 52.59 & \textbf{59.93} & +6.05 & +7.33 \\
& Inst Loose    & 85.44 & 83.23 & 82.76 & \textbf{85.87} & +2.64 & +3.11 \\
\midrule
\multicolumn{2}{l}{Turn 1 Average} & 85.72 & 86.94 & 87.21 & \textbf{87.93} & +0.99 & +0.72 \\
\multicolumn{2}{l}{Turn 2 Average} & 78.07 & 76.06 & 75.78 & \textbf{79.15} & +3.09 & +3.37 \\
\multicolumn{2}{l}{Turn 3 Average} & 69.45 & 66.74 & 66.14 & \textbf{70.48} & +3.74 & +4.34 \\
\midrule
\multicolumn{2}{l}{Overall Average} & 77.75 & 76.58 & 76.38 & \textbf{79.19} & +2.61 & +2.81 \\
\bottomrule
\end{tabular}
}
\caption{Full Multi-IF results with an early Forward-DPO checkpoint. All scores are macro-averaged across eight languages (Chinese, English, Italian, Spanish, Hindi, Portuguese, French, Russian). Fwd-60 is only marginally above Fwd-100 on Multi-IF overall (+0.20) and remains below both Base and BPS. BPS is above Fwd-60 on 11 of 12 individual metrics, and the gap widens with turn depth, showing that the later-turn separation persists at the tested earlier checkpoint.}
\label{tab:multiif_full}
\end{table*}

\section{Per-Language Multi-IF Analysis: Depth-Dependent Degradation and BPS Recovery}
\label{sec:multiif_language}

Table~\ref{tab:multiif_language_deltas} decomposes the Multi-IF deltas by language and turn depth.

\begin{table*}[!t]
\centering
{
\small
\begin{tabular}{lrrr|rrr|rrr}
\toprule
\multirow{2}{*}{\textbf{Language}}
& \multicolumn{3}{c|}{$\Delta$ \textbf{Fwd$-$Base}}
& \multicolumn{3}{c|}{$\Delta$ \textbf{BPS$-$Base}}
& \multicolumn{3}{c}{$\Delta$ \textbf{BPS$-$Fwd}} \\
\cmidrule(lr){2-4}\cmidrule(lr){5-7}\cmidrule(lr){8-10}
& T1 & T2 & T3 & T1 & T2 & T3 & T1 & T2 & T3 \\
\midrule
Chinese    & $+$1.00 & $-$1.86 & $-$1.69 & $+$1.20 & $-$0.04 & $+$1.30 & $+$0.20 & $+$1.82 & $+$2.99 \\
English    & $+$0.92 & $-$2.21 & $-$4.28 & $+$1.89 & $+$0.04 & $+$0.29 & $+$0.97 & $+$2.25 & $+$4.58 \\
Italian    & $+$1.87 & $+$0.02 & $-$2.25 & $+$2.40 & $+$1.28 & $+$0.63 & $+$0.53 & $+$1.26 & $+$2.89 \\
Spanish    & $+$1.67 & $-$3.17 & $-$3.85 & $+$3.09 & $+$0.77 & $+$0.49 & $+$1.43 & $+$3.94 & $+$4.34 \\
Hindi      & $+$2.68 & $-$1.62 & $-$0.83 & $+$2.07 & $+$1.72 & $+$2.91 & $-$0.61 & $+$3.33 & $+$3.74 \\
Portuguese & $+$0.58 & $-$2.37 & $-$4.11 & $+$1.14 & $+$1.76 & $+$0.94 & $+$0.56 & $+$4.13 & $+$5.05 \\
French     & $+$1.91 & $-$0.99 & $-$4.20 & $+$2.08 & $+$1.74 & $+$0.15 & $+$0.17 & $+$2.74 & $+$4.35 \\
Russian    & $+$1.31 & $-$6.07 & $-$5.33 & $+$3.80 & $+$1.40 & $+$1.49 & $+$2.49 & $+$7.47 & $+$6.82 \\
\midrule
\textbf{Macro Avg.} & $+$1.49 & $-$2.29 & $-$3.32 & $+$2.21 & $+$1.08 & $+$1.03 & $+$0.72 & $+$3.37 & $+$4.34 \\
\bottomrule
\end{tabular}
}
\caption{Per-language Multi-IF turn-level deltas in percentage points. For each language and turn, the score averages the four Multi-IF metrics (Prompt Strict, Inst Strict, Prompt Loose, Inst Loose); deltas compare against this per-language turn average. Fwd denotes Fwd-100; Macro Avg.\ averages languages uniformly. Forward-DPO improves turn-1 in all 8 languages but falls below Base in turn-2 for 7/8 languages and in turn-3 for 8/8 languages. BPS exceeds Fwd-100 in 23/24 language-turn averages. At the finer language$\times$metric level, BPS beats Fwd-100 in all 64 cells from turns 2--3 and in 85/96 cells overall, with 2 additional ties.}
\label{tab:multiif_language_deltas}
\end{table*}

\section{Output Length Analysis}
\label{sec:length_appendix}

Table~\ref{tab:length} compares average output token lengths across models and datasets.

\begin{table}[t]
\centering
{
\small
\begin{tabular}{llr}
\toprule
\textbf{Dataset} & \textbf{Model} & \textbf{Avg.\ Output Tokens} \\
\midrule
\multirow{3}{*}{IFBench}  & Base               & 1{,}584.03 \\
                           & Forward-DPO (S100) & 5{,}919.98 \\
                           & BPS (S100)         & 2{,}564.96 \\
\midrule
\multirow{3}{*}{ToolBench} & Base               & 165.58 \\
                            & Forward-DPO (S100) & 205.17 \\
                            & BPS (S100)         & 157.30 \\
\bottomrule
\end{tabular}
}
\caption{Average output tokens per response. Forward-DPO outputs are 3.7$\times$ longer than Base on IFBench and are also longer on ToolBench, alongside lower performance on several reported metrics. BPS maintains generation lengths much closer to Base.}
\label{tab:length}
\end{table}

\section{Additional Model and Seed-Data Checks}
\label{sec:generalization_full}

Table~\ref{tab:cross_model} reports additional checks on Qwen3-4B (base checkpoint) and Qwen2.5-7B-Instruct using IFEval prompts as seeds.
\begin{table*}[!t]
\centering
{
\footnotesize
\begin{tabular}{lcccccc}
\toprule
\multirow{2}{*}{\textbf{Benchmark}} & \multicolumn{3}{c}{\textbf{Qwen2.5-7B-Instruct}~\cite{qwen2025qwen25technicalreport}} & \multicolumn{3}{c}{\textbf{Qwen3-4B}} \\
\cmidrule(lr){2-4}\cmidrule(lr){5-7}
 & Base & Fwd & BPS & Base & Fwd & BPS \\
\midrule
\multicolumn{7}{l}{\textit{Instruction Following (IFBench)}} \\
\quad Strict-P & 25.33 & 31.67 & \textbf{33.00} & 24.00 & 25.00 & \textbf{27.30} \\
\quad Strict-I & 27.03 & 32.85 & \textbf{35.17} & 25.90 & 27.60 & \textbf{29.10} \\
\quad Loose-P  & 29.33 & 34.33 & \textbf{37.67} & 28.00 & 28.30 & \textbf{30.70} \\
\quad Loose-I  & 31.69 & 36.63 & \textbf{40.70} & 29.40 & 31.70 & \textbf{33.10} \\
\midrule
\multicolumn{7}{l}{\textit{Code Generation}} \\
\quad HumanEval  & 68.90 & 68.90 & \textbf{71.34} & 71.95 & 68.90 & \textbf{74.39} \\
\quad HumanEval+ & ---   & ---   & ---            & 64.02 & 65.85 & \textbf{68.29} \\
\midrule
\multicolumn{7}{l}{\textit{General Capability}} \\
\quad MMLU       & \textbf{74.31} & 74.26 & 74.13 & 70.05 & 70.00 & \textbf{70.07} \\
\quad TruthfulQA & ---   & ---   & ---            & 40.76 & 48.84 & \textbf{50.55} \\
\bottomrule
\end{tabular}
}
\caption{Additional model and seed-data checks on Qwen3-4B (the base checkpoint, distinct from the main Qwen3-4B-Instruct-2507) and Qwen2.5-7B-Instruct using IFEval prompts as seeds. BPS outperforms Forward-DPO on all IFBench metrics in these two settings (+1.40 to +4.07 points), and improves HumanEval while Forward-DPO stagnates or degrades. MMLU remains stable across all settings.}
\label{tab:cross_model}
\end{table*}

\section{Failure Type Breakdown}
\label{sec:failure_type_appendix}

We classify validated reverse anchors by the primary type of boundary failure. Because IFBench-style prompts often combine surface constraints with task content, this analysis is descriptive rather than a claim about the full distribution of all possible boundary failures.

\begin{table*}[t]
\centering
{
\footnotesize
\begin{tabular}{lrrrrr}
\toprule
\textbf{Type} & \textbf{Gemini} & \textbf{Kimi} & \textbf{Total} & \textbf{Held-out} & \textbf{Total \%} \\
\midrule
Format/structure mismatch   & 153 & 82 & 235 & 91 & 54.3 \\
Multi-constraint mixed      &  53 & 36 &  89 & 53 & 20.6 \\
Lexical-constraint mismatch &  63 & 23 &  86 & 37 & 19.9 \\
Content-intent mismatch     &   8 &  9 &  17 & 10 &  3.9 \\
Language mismatch           &   5 &  1 &   6 &  4 &  1.4 \\
\midrule
\textbf{Total}              & 282 & 151 & 433 & 195 & 100.0 \\
\bottomrule
\end{tabular}
}
\caption{Failure-type distribution among validated reverse anchors. Format/structure and lexical categories are common because IFBench prompts contain explicit surface constraints such as paragraph counts, capitalization, start/end phrases, required keywords, and separators.}
\label{tab:failure_type_distribution}
\end{table*}

\begin{table*}[t]
\centering
{
\footnotesize
\begin{tabular}{lrrrrrrr}
\toprule
\textbf{Type} & \textbf{N} & \textbf{Fwd $M_\text{ach}$} & \textbf{BPS $M_\text{ach}$} & $\Delta M$ & \textbf{Fwd Acc.} & \textbf{BPS Acc.} & $\Delta$ Acc. \\
\midrule
Content-intent mismatch     & 10 & \phantom{$-$}0.115 & \phantom{$-$}0.144 & +0.030 & 60.0 & 100.0 & +40.0 \\
Language mismatch           &  4 & $-$0.100 & \phantom{$-$}0.022 & +0.122 &  0.0 &  50.0 & +50.0 \\
Format/structure mismatch   & 91 & $-$0.411 & \phantom{$-$}0.033 & +0.444 &  2.2 &  71.4 & +69.2 \\
Lexical-constraint mismatch & 37 & $-$0.362 & $-$0.005 & +0.357 &  5.4 &  67.6 & +62.2 \\
Multi-constraint mixed      & 53 & $-$0.217 & \phantom{$-$}0.049 & +0.267 & 11.3 &  62.3 & +50.9 \\
\bottomrule
\end{tabular}
}
\caption{Held-out achieved-prompt reward geometry by failure type. BPS raises achieved-side pairwise ranking accuracy for every observed type. The largest gains occur on format/structure and lexical-constraint mismatches, the dominant IFBench-derived boundary-failure categories. Small categories such as content-intent and language mismatch are included for completeness.}
\label{tab:failure_type_reward}
\end{table*}

Tables~\ref{tab:failure_type_distribution} and~\ref{tab:failure_type_reward} show that BPS does not require content-level semantic swaps to be effective. Most validated reverse anchors are surface-form boundary failures, where the failed response is coherent but violates constraints such as structure, casing, required terms, or separators. Forward-DPO rarely assigns the intended achieved-side ordering in the main categories, while BPS substantially raises the corresponding pairwise ranking accuracy. This supports the view that BPS retains response modes associated with constrained instruction following, not only alternative factual intents.

\section{Three-Seed Reward-Geometry Stability}
\label{sec:seed_stability_appendix}

To test whether the crossed-anchor reward-geometry result depends on a single training seed, we train Forward-DPO and BPS with three seeds under the same data, hyperparameters, and fixed step-100 checkpoint rule. Table~\ref{tab:seed_stability} reports the Gemini held-out crossed-anchor probe.

\begin{table*}[t]
\centering
{
\footnotesize
\begin{tabular}{lrrrr}
\toprule
\textbf{Model / Seed} & $M_\text{orig}$ & Acc$_\text{orig}$ & $M_\text{achieved}$ & Acc$_\text{achieved}$ \\
\midrule
Fwd 42 & 0.542 & 99.3 & $-$0.351 &  7.5 \\
Fwd 43 & 0.512 & 99.3 & $-$0.338 &  8.9 \\
Fwd 44 & 0.527 & 99.3 & $-$0.337 &  7.5 \\
\midrule
Fwd mean$\pm$std & 0.527$\pm$0.013 & 99.3$\pm$0.0 & $-$0.342$\pm$0.007 & 8.0$\pm$0.6 \\
\midrule
BPS 42 & 0.299 & 100.0 & \phantom{$-$}0.007 & 63.7 \\
BPS 43 & 0.293 & 100.0 & \phantom{$-$}0.015 & 63.7 \\
BPS 44 & 0.284 &  99.3 & \phantom{$-$}0.031 & 69.2 \\
\midrule
BPS mean$\pm$std & 0.292$\pm$0.006 & 99.8$\pm$0.3 & \phantom{$-$}0.018$\pm$0.010 & 65.5$\pm$2.6 \\
\bottomrule
\end{tabular}
}
\caption{Three-seed stability on the Gemini held-out crossed-anchor reward probe. BPS consistently preserves original-side pairwise ranking and raises achieved-side ranking accuracy across seeds. The pooled paired achieved-margin improvement is $+0.360$ (95\% bootstrap CI $[0.320,0.401]$).}
\label{tab:seed_stability}
\end{table*}

The achieved-side gap is stable across seeds: Forward-DPO remains near 8\% pairwise ranking accuracy, while BPS remains above 63\% for all three runs. Original-side ranking remains saturated for both methods, so the achieved-side shift is obtained without sacrificing the forward preference boundary.

\section{Three-Seed IFEval Stability}
\label{sec:ifeval_seed_appendix}

Table~\ref{tab:ifeval_seed_stability} reports IFEval results over three training seeds. The gains are smaller than the crossed-anchor and Multi-IF effects, but the average IFEval score is consistently higher for BPS.

\begin{table*}[t]
\centering
{
\footnotesize
\begin{tabular}{lrrrrr}
\toprule
\textbf{Model / Seed} & \textbf{P Strict} & \textbf{P Loose} & \textbf{I Strict} & \textbf{I Loose} & \textbf{Avg.} \\
\midrule
Fwd 42 & 83.55 & 85.21 & 90.36 & 91.34 & 87.62 \\
Fwd 43 & 83.92 & 85.21 & 90.42 & 91.13 & 87.67 \\
Fwd 44 & 84.10 & 85.77 & 90.76 & 91.71 & 88.09 \\
\midrule
Fwd mean$\pm$std & 83.86$\pm$0.23 & 85.40$\pm$0.26 & 90.51$\pm$0.18 & 91.39$\pm$0.24 & 87.79$\pm$0.23 \\
\midrule
BPS 42 & 83.73 & 87.25 & 89.49 & 91.74 & 88.05 \\
BPS 43 & 84.84 & 88.35 & 90.33 & 92.30 & 88.96 \\
BPS 44 & 85.40 & 88.72 & 90.88 & 92.88 & 89.47 \\
\midrule
BPS mean$\pm$std & 84.66$\pm$0.69 & 88.11$\pm$0.62 & 90.23$\pm$0.57 & 92.31$\pm$0.47 & 88.83$\pm$0.59 \\
\bottomrule
\end{tabular}
}
\caption{Three-seed IFEval stability at the fixed step-100 comparison. Scores are percentages. BPS improves prompt-level strict, prompt-level loose, and instruction-level loose metrics on average, while instruction-level strict is slightly lower. The overall IFEval average improves from 87.79$\pm$0.23 to 88.83$\pm$0.59.}
\label{tab:ifeval_seed_stability}
\end{table*}

\section{Human Audit of Achieved Prompts}
\label{sec:human_audit}

We conduct a blind audit of 200 held-out crossed anchors, split evenly between the Gemini-source and Kimi-source probes. Each anchor is judged by three annotators (600 total judgments). For each anchor, annotators are shown the achieved prompt and two randomly ordered responses, without seeing the teacher label or whether a response is $\tau_\text{fail}$ or $\tau_\text{perfect}$. They judge whether the achieved prompt is fully natural, whether $\tau_\text{fail}$ satisfies it, which response is preferred under the achieved prompt, and whether the prompt leaks response content. Table~\ref{tab:human_audit} reports per-anchor majority-vote results.

\begin{table*}[t]
\centering
{
\footnotesize
\begin{tabular}{lrrrrrr}
\toprule
\textbf{Source} & \textbf{N} & \textbf{Natural} & \textbf{Fail sat.} & \textbf{Rev. pref.} & \textbf{Leakage}$\downarrow$ & \textbf{Accepted} \\
\midrule
Gemini & 100 & 99.0 & 100.0 & 100.0 & 0.0 & 99.0 \\
Kimi   & 100 & 83.0 &  87.0 &  98.0 & 1.0 & 67.0 \\
All    & 200 & 91.0 &  93.5 &  99.0 & 0.5 & 83.0 \\
\bottomrule
\end{tabular}
}
\caption{Blind human audit of achieved prompts and reverse preferences. Values are per-anchor majority-vote percentages over three annotators per anchor. Accepted means that the achieved prompt is natural, $\tau_\text{fail}$ satisfies it, the reverse preference is correct, and no leakage is flagged.}
\label{tab:human_audit}
\end{table*}

Table~\ref{tab:human_audit} shows that the central supervision signal---the achieved-prompt preference direction---is highly reliable under blind audit: annotators prefer $\tau_\text{fail}$ under $P_\text{achieved}$ in 99.0\% of audited anchors, leakage is rare (0.5\%), and 91.0\% of achieved prompts are judged natural user requests by majority vote. At the annotator-judgment level, naturalness is 87.2\% overall (523/600), with 95.7\% for Gemini-source prompts and 78.7\% for Kimi-source prompts. The stricter joint acceptance measure is necessarily lower because it requires all four criteria simultaneously; importantly, reverse-preference agreement remains 100\% for Gemini-source and 98\% for Kimi-source anchors. Prompts judged less natural typically inherit explicit surface constraints such as exact paragraph counts, required keywords, forbidden words, fixed capitalization, or prescribed separators, rather than leaking response content. Thus, the audit strongly supports the reverse preference direction and indicates that most achieved prompts remain plausible standalone requests.

\section{Teacher Prompts and Validation Rubric}
\label{sec:teacher_prompts}

All teacher calls use Gemini-3.1-pro-preview. Below are simplified descriptions of the three prompt templates used in BPS.

\paragraph{Correction hint prompt.}
The teacher receives the original prompt, the failed response, and the list of unsatisfied constraints. It is asked to produce a structured hint identifying: (1) which constraints were violated and why; (2) what specific changes are needed; (3) which factual content from the failed response can be retained. The student then produces $\tau_\text{perfect}$ conditioned on this hint.

\paragraph{Achieved prompt generation prompt.}
For reverse-eligible failures, the teacher receives the original prompt, the failed response, the outcome summary, and an eligibility judgment. If eligible, it is asked to generate a standalone user-facing achieved prompt $P_\text{achieved}$ satisfying: (1) phrased as a user request without quoting the failed response; (2) explicitly encodes the language, format, and structural properties of $\tau_\text{fail}$ as requirements; (3) does not request false or hallucinated content; (4) does not implicitly ask the model to reproduce $\tau_\text{fail}$.

\paragraph{Reverse pair validation rubric.}
A judge accepts the reverse pair $(P_\text{achieved}, \tau_\text{fail} \succ \tau_\text{perfect})$ only if \emph{all} of the following hold:
\begin{enumerate}
\item $P_\text{achieved}$ reads as a standalone user prompt.
\item $\tau_\text{fail}$ genuinely and completely satisfies $P_\text{achieved}$.
\item $\tau_\text{fail}$ is coherent, truthful, and safe.
\item $\tau_\text{perfect}$ is clearly worse under $P_\text{achieved}$ (e.g., wrong language, incompatible format).
\item $P_\text{achieved}$ does not implicitly ask to reproduce or copy $\tau_\text{fail}$.
\end{enumerate}
Pairs where both responses satisfy $P_\text{achieved}$, or where the preference direction is ambiguous, are discarded.

\section{Checkpoint Sensitivity of Forward-DPO}
\label{sec:fwd_checkpoint_appendix}

Table~\ref{tab:fwd_checkpoint_sensitivity} compares the early Fwd-60 checkpoint against the matched Fwd-100 and BPS-100 checkpoints across instruction-following, code, and agentic/tool-use metrics.

\begin{table*}[t]
\centering
{
\small
\begin{tabular}{lccccc}
\toprule
\textbf{Metric} & \textbf{Base} & \textbf{Fwd-60} & \textbf{Fwd-100} & \textbf{BPS-100} & $\Delta_{\text{BPS-F60}}$ \\
\midrule
IFBench Avg.          & 33.63 & \textbf{36.22} & 34.39 & 35.46 & $-$0.76 \\
IFEval Avg.           & 87.18 & 88.46 & 87.36 & \textbf{90.27} & +1.81 \\
\midrule
Multi-IF Turn 1 Avg.  & 85.72 & 86.94 & 87.21 & \textbf{87.93} & +0.99 \\
Multi-IF Turn 2 Avg.  & 78.07 & 76.06 & 75.78 & \textbf{79.15} & +3.09 \\
Multi-IF Turn 3 Avg.  & 69.45 & 66.74 & 66.14 & \textbf{70.48} & +3.74 \\
Multi-IF Overall Avg. & 77.75 & 76.58 & 76.38 & \textbf{79.19} & +2.61 \\
\midrule
LiveCodeBench         & \textbf{48.91} & 45.40 & 45.40 & 48.53 & +3.13 \\
MBPP                  & 72.00 & 70.60 & 72.00 & \textbf{72.80} & +2.20 \\
tau-bench airline     & \textbf{36.00} & 28.00 & 30.00 & \textbf{36.00} & +8.00 \\
tau2-bench airline    & \textbf{27.91} & 25.58 & 25.58 & \textbf{27.91} & +2.33 \\
tau2-bench retail     & \textbf{43.75} & 38.89 & 36.61 & \textbf{43.75} & +4.86 \\
ToolBench Act.EM (ov.)& \textbf{35.50} & 33.58 & 33.10 & 35.46 & +1.88 \\
ToolBench OOD F1      & \textbf{29.38} & 25.99 & 26.75 & 28.98 & +2.99 \\
ToolBench HalluRate (OOD)$\downarrow$ & \textbf{0.53} & 0.79 & 0.66 & \textbf{0.53} & $-$0.26 \\
\bottomrule
\end{tabular}
}
\caption{Forward-DPO checkpoint sensitivity. Fwd-60 has a higher narrow correction-oriented IFBench average than Fwd-100, but remains below BPS on the listed multilingual multi-turn, code, agentic, and tool-use checks. The main paper uses the common step-100 checkpoint to match optimizer updates across methods; this earlier checkpoint confirms that the broader capability-retention pattern is not specific to the step-100 Forward-DPO checkpoint. For HalluRate, lower is better and the last column reports BPS$-$Fwd-60.}
\label{tab:fwd_checkpoint_sensitivity}
\end{table*}

\section{Training-Anchor Reward Probe}
A probe on the 282 training reverse anchors shows the same pattern: BPS raises achieved-side pairwise ranking accuracy from 7.1\% to 75.9\% and paired achieved margins by +0.44 (95\% CI $[0.36,0.54]$).

\bibliography{custom}

@article{rafailov2023direct,
  title={Direct preference optimization: Your language model is secretly a reward model},
  author={Rafailov, Rafael and Sharma, Archit and Mitchell, Eric and Manning, Christopher D and Ermon, Stefano and Finn, Chelsea},
  journal={Advances in neural information processing systems},
  volume={36},
  pages={53728--53741},
  year={2023}
}

@inproceedings{azar2024general,
  title={A general theoretical paradigm to understand learning from human preferences},
  author={Azar, Mohammad Gheshlaghi and Guo, Zhaohan Daniel and Piot, Bilal and Munos, Remi and Rowland, Mark and Valko, Michal and Calandriello, Daniele},
  booktitle={International Conference on Artificial Intelligence and Statistics},
  pages={4447--4455},
  year={2024},
  organization={PMLR}
}

@misc{ethayarajh2024ktomodelalignmentprospect,
      title={KTO: Model Alignment as Prospect Theoretic Optimization}, 
      author={Kawin Ethayarajh and Winnie Xu and Niklas Muennighoff and Dan Jurafsky and Douwe Kiela},
      year={2024},
      eprint={2402.01306},
      archivePrefix={arXiv},
      primaryClass={cs.LG},
      url={https://arxiv.org/abs/2402.01306}, 
}

@article{meng2024simpo,
  title={Simpo: Simple preference optimization with a reference-free reward},
  author={Meng, Yu and Xia, Mengzhou and Chen, Danqi},
  journal={Advances in Neural Information Processing Systems},
  volume={37},
  pages={124198--124235},
  year={2024}
}

@inproceedings{hong2024orpo,
  title={Orpo: Monolithic preference optimization without reference model},
  author={Hong, Jiwoo and Lee, Noah and Thorne, James},
  booktitle={Proceedings of the 2024 Conference on Empirical Methods in Natural Language Processing},
  pages={11170--11189},
  year={2024}
}

@article{zelikman2022star,
  title={Star: Bootstrapping reasoning with reasoning},
  author={Zelikman, Eric and Wu, Yuhuai and Mu, Jesse and Goodman, Noah},
  journal={Advances in Neural Information Processing Systems},
  volume={35},
  pages={15476--15488},
  year={2022}
}

@article{madaan2023self,
  title={Self-refine: Iterative refinement with self-feedback},
  author={Madaan, Aman and Tandon, Niket and Gupta, Prakhar and Hallinan, Skyler and Gao, Luyu and Wiegreffe, Sarah and Alon, Uri and Dziri, Nouha and Prabhumoye, Shrimai and Yang, Yiming and others},
  journal={Advances in neural information processing systems},
  volume={36},
  pages={46534--46594},
  year={2023}
}

@misc{touvron2023llama2openfoundation,
      title={Llama 2: Open Foundation and Fine-Tuned Chat Models}, 
      author={Hugo Touvron and Louis Martin and Kevin Stone and Peter Albert and Amjad Almahairi and Yasmine Babaei and Nikolay Bashlykov and Soumya Batra and Prajjwal Bhargava and Shruti Bhosale and Dan Bikel and Lukas Blecher and Cristian Canton Ferrer and Moya Chen and Guillem Cucurull and David Esiobu and Jude Fernandes and Jeremy Fu and Wenyin Fu and Brian Fuller and Cynthia Gao and Vedanuj Goswami and Naman Goyal and Anthony Hartshorn and Saghar Hosseini and Rui Hou and Hakan Inan and Marcin Kardas and Viktor Kerkez and Madian Khabsa and Isabel Kloumann and Artem Korenev and Punit Singh Koura and Marie-Anne Lachaux and Thibaut Lavril and Jenya Lee and Diana Liskovich and Yinghai Lu and Yuning Mao and Xavier Martinet and Todor Mihaylov and Pushkar Mishra and Igor Molybog and Yixin Nie and Andrew Poulton and Jeremy Reizenstein and Rashi Rungta and Kalyan Saladi and Alan Schelten and Ruan Silva and Eric Michael Smith and Ranjan Subramanian and Xiaoqing Ellen Tan and Binh Tang and Ross Taylor and Adina Williams and Jian Xiang Kuan and Puxin Xu and Zheng Yan and Iliyan Zarov and Yuchen Zhang and Angela Fan and Melanie Kambadur and Sharan Narang and Aurelien Rodriguez and Robert Stojnic and Sergey Edunov and Thomas Scialom},
      year={2023},
      eprint={2307.09288},
      archivePrefix={arXiv},
      primaryClass={cs.CL},
      url={https://arxiv.org/abs/2307.09288}, 
}

@misc{pyatkin2025generalizingverifiableinstructionfollowing,
      title={Generalizing Verifiable Instruction Following}, 
      author={Valentina Pyatkin and Saumya Malik and Victoria Graf and Hamish Ivison and Shengyi Huang and Pradeep Dasigi and Nathan Lambert and Hannaneh Hajishirzi},
      year={2025},
      eprint={2507.02833},
      archivePrefix={arXiv},
      primaryClass={cs.CL},
      url={https://arxiv.org/abs/2507.02833}, 
}

@inproceedings{patil2025berkeley,
  title={The berkeley function calling leaderboard (bfcl): From tool use to agentic evaluation of large language models},
  author={Patil, Shishir G and Mao, Huanzhi and Yan, Fanjia and Ji, Charlie Cheng-Jie and Suresh, Vishnu and Stoica, Ion and Gonzalez, Joseph E},
  booktitle={Forty-second International Conference on Machine Learning},
  year={2025}
}

@misc{yao2024taubenchbenchmarktoolagentuserinteraction,
      title={$\tau$-bench: A Benchmark for Tool-Agent-User Interaction in Real-World Domains}, 
      author={Shunyu Yao and Noah Shinn and Pedram Razavi and Karthik Narasimhan},
      year={2024},
      eprint={2406.12045},
      archivePrefix={arXiv},
      primaryClass={cs.AI},
      url={https://arxiv.org/abs/2406.12045}, 
}

@inproceedings{jiang2024followbench,
  title={Followbench: A multi-level fine-grained constraints following benchmark for large language models},
  author={Jiang, Yuxin and Wang, Yufei and Zeng, Xingshan and Zhong, Wanjun and Li, Liangyou and Mi, Fei and Shang, Lifeng and Jiang, Xin and Liu, Qun and Wang, Wei},
  booktitle={Proceedings of the 62nd Annual Meeting of the Association for Computational Linguistics (Volume 1: Long Papers)},
  pages={4667--4688},
  year={2024}
}

@misc{qin2023toolllmfacilitatinglargelanguage,
      title={ToolLLM: Facilitating Large Language Models to Master 16000+ Real-world APIs}, 
      author={Yujia Qin and Shihao Liang and Yining Ye and Kunlun Zhu and Lan Yan and Yaxi Lu and Yankai Lin and Xin Cong and Xiangru Tang and Bill Qian and Sihan Zhao and Lauren Hong and Runchu Tian and Ruobing Xie and Jie Zhou and Mark Gerstein and Dahai Li and Zhiyuan Liu and Maosong Sun},
      year={2023},
      eprint={2307.16789},
      archivePrefix={arXiv},
      primaryClass={cs.AI},
      url={https://arxiv.org/abs/2307.16789}, 
}

@article{andrychowicz2017hindsight,
  title={Hindsight experience replay},
  author={Andrychowicz, Marcin and Wolski, Filip and Ray, Alex and Schneider, Jonas and Fong, Rachel and Welinder, Peter and McGrew, Bob and Tobin, Josh and Pieter Abbeel, OpenAI and Zaremba, Wojciech},
  journal={Advances in neural information processing systems},
  volume={30},
  year={2017}
}

@inproceedings{huang2025reverse,
  title={Reverse preference optimization for complex instruction following},
  author={Huang, Xiang and Lin, Ting-En and Fang, Feiteng and Wu, Yuchuan and Li, Hangyu and Qu, Yuzhong and Huang, Fei and Li, Yongbin},
  booktitle={Findings of the Association for Computational Linguistics: ACL 2025},
  pages={20700--20723},
  year={2025}
}

@inproceedings{zhang2025iopo,
  title={Iopo: Empowering llms with complex instruction following via input-output preference optimization},
  author={Zhang, Xinghua and Yu, Haiyang and Fu, Cheng and Huang, Fei and Li, Yongbin},
  booktitle={Proceedings of the 63rd Annual Meeting of the Association for Computational Linguistics (Volume 1: Long Papers)},
  pages={22185--22200},
  year={2025}
}

@inproceedings{zhang2023wisdom,
  title={The wisdom of hindsight makes language models better instruction followers},
  author={Zhang, Tianjun and Liu, Fangchen and Wong, Justin and Abbeel, Pieter and Gonzalez, Joseph E},
  booktitle={International Conference on Machine Learning},
  pages={41414--41428},
  year={2023},
  organization={PMLR}
}

@inproceedings{razin2025unintentional,
  title={Unintentional unalignment: Likelihood displacement in direct preference optimization},
  author={Razin, Noam and Malladi, Sadhika and Bhaskar, Adithya and Chen, Danqi and Arora, Sanjeev and Hanin, Boris},
  booktitle={International Conference on Learning Representations},
  volume={2025},
  pages={24791--24834},
  year={2025}
}

@inproceedings{zhang2025xlam,
  title={xlam: A family of large action models to empower ai agent systems},
  author={Zhang, Jianguo and Lan, Tian and Zhu, Ming and Liu, Zuxin and Hoang, Thai Quoc and Kokane, Shirley and Yao, Weiran and Tan, Juntao and Prabhakar, Akshara and Chen, Haolin and others},
  booktitle={Proceedings of the 2025 Conference of the Nations of the Americas Chapter of the Association for Computational Linguistics: Human Language Technologies (Volume 1: Long Papers)},
  pages={11583--11597},
  year={2025}
}

@misc{qwen2025qwen25technicalreport,
      title={Qwen2.5 Technical Report}, 
      author={Qwen and : and An Yang and Baosong Yang and Beichen Zhang and Binyuan Hui and Bo Zheng and Bowen Yu and Chengyuan Li and Dayiheng Liu and Fei Huang and Haoran Wei and Huan Lin and Jian Yang and Jianhong Tu and Jianwei Zhang and Jianxin Yang and Jiaxi Yang and Jingren Zhou and Junyang Lin and Kai Dang and Keming Lu and Keqin Bao and Kexin Yang and Le Yu and Mei Li and Mingfeng Xue and Pei Zhang and Qin Zhu and Rui Men and Runji Lin and Tianhao Li and Tianyi Tang and Tingyu Xia and Xingzhang Ren and Xuancheng Ren and Yang Fan and Yang Su and Yichang Zhang and Yu Wan and Yuqiong Liu and Zeyu Cui and Zhenru Zhang and Zihan Qiu},
      year={2025},
      eprint={2412.15115},
      archivePrefix={arXiv},
      primaryClass={cs.CL},
      url={https://arxiv.org/abs/2412.15115}, 
}

@inproceedings{rasley2020deepspeed,
  title={Deepspeed: System optimizations enable training deep learning models with over 100 billion parameters},
  author={Rasley, Jeff and Rajbhandari, Samyam and Ruwase, Olatunji and He, Yuxiong},
  booktitle={Proceedings of the 26th ACM SIGKDD international conference on knowledge discovery \& data mining},
  pages={3505--3506},
  year={2020}
}

@article{liu2023your,
  title={Is your code generated by chatgpt really correct? rigorous evaluation of large language models for code generation},
  author={Liu, Jiawei and Xia, Chunqiu Steven and Wang, Yuyao and Zhang, Lingming},
  journal={Advances in neural information processing systems},
  volume={36},
  pages={21558--21572},
  year={2023}
}

@article{huang2023c,
  title={C-eval: A multi-level multi-discipline chinese evaluation suite for foundation models},
  author={Huang, Yuzhen and Bai, Yuzhuo and Zhu, Zhihao and Zhang, Junlei and Zhang, Jinghan and Su, Tangjun and Liu, Junteng and Lv, Chuancheng and Zhang, Yikai and Fu, Yao and others},
  journal={Advances in neural information processing systems},
  volume={36},
  pages={62991--63010},
  year={2023}
}

@article{wang2024mmlu,
  title={Mmlu-pro: A more robust and challenging multi-task language understanding benchmark},
  author={Wang, Yubo and Ma, Xueguang and Zhang, Ge and Ni, Yuansheng and Chandra, Abhranil and Guo, Shiguang and Ren, Weiming and Arulraj, Aaran and He, Xuan and Jiang, Ziyan and others},
  journal={Advances in Neural Information Processing Systems},
  volume={37},
  pages={95266--95290},
  year={2024}
}

@inproceedings{suzgun2023challenging,
  title={Challenging big-bench tasks and whether chain-of-thought can solve them},
  author={Suzgun, Mirac and Scales, Nathan and Sch{\"a}rli, Nathanael and Gehrmann, Sebastian and Tay, Yi and Chung, Hyung Won and Chowdhery, Aakanksha and Le, Quoc and Chi, Ed H and Zhou, Denny and others},
  booktitle={Findings of the Association for Computational Linguistics: ACL 2023},
  pages={13003--13051},
  year={2023}
}

@misc{barres2025tau2benchevaluatingconversationalagents,
      title={$\tau^2$-Bench: Evaluating Conversational Agents in a Dual-Control Environment}, 
      author={Victor Barres and Honghua Dong and Soham Ray and Xujie Si and Karthik Narasimhan},
      year={2025},
      eprint={2506.07982},
      archivePrefix={arXiv},
      primaryClass={cs.AI},
      url={https://arxiv.org/abs/2506.07982}, 
}

@article{christiano2017deep,
  title={Deep reinforcement learning from human preferences},
  author={Christiano, Paul F and Leike, Jan and Brown, Tom and Martic, Miljan and Legg, Shane and Amodei, Dario},
  journal={Advances in neural information processing systems},
  volume={30},
  year={2017}
}

@article{stiennon2020learning,
  title={Learning to summarize with human feedback},
  author={Stiennon, Nisan and Ouyang, Long and Wu, Jeffrey and Ziegler, Daniel and Lowe, Ryan and Voss, Chelsea and Radford, Alec and Amodei, Dario and Christiano, Paul F},
  journal={Advances in neural information processing systems},
  volume={33},
  pages={3008--3021},
  year={2020}
}

@article{ouyang2022training,
  title={Training language models to follow instructions with human feedback},
  author={Ouyang, Long and Wu, Jeffrey and Jiang, Xu and Almeida, Diogo and Wainwright, Carroll and Mishkin, Pamela and Zhang, Chong and Agarwal, Sandhini and Slama, Katarina and Ray, Alex and others},
  journal={Advances in neural information processing systems},
  volume={35},
  pages={27730--27744},
  year={2022}
}

@misc{lee2024rlaifvsrlhfscaling,
      title={RLAIF vs. RLHF: Scaling Reinforcement Learning from Human Feedback with AI Feedback}, 
      author={Harrison Lee and Samrat Phatale and Hassan Mansoor and Thomas Mesnard and Johan Ferret and Kellie Lu and Colton Bishop and Ethan Hall and Victor Carbune and Abhinav Rastogi and Sushant Prakash},
      year={2024},
      eprint={2309.00267},
      archivePrefix={arXiv},
      primaryClass={cs.CL},
      url={https://arxiv.org/abs/2309.00267}, 
}

@misc{zhao2023slichfsequencelikelihoodcalibration,
      title={SLiC-HF: Sequence Likelihood Calibration with Human Feedback}, 
      author={Yao Zhao and Rishabh Joshi and Tianqi Liu and Misha Khalman and Mohammad Saleh and Peter J. Liu},
      year={2023},
      eprint={2305.10425},
      archivePrefix={arXiv},
      primaryClass={cs.CL},
      url={https://arxiv.org/abs/2305.10425}, 
}

@misc{yuan2023rrhfrankresponsesalign,
      title={RRHF: Rank Responses to Align Language Models with Human Feedback without tears}, 
      author={Zheng Yuan and Hongyi Yuan and Chuanqi Tan and Wei Wang and Songfang Huang and Fei Huang},
      year={2023},
      eprint={2304.05302},
      archivePrefix={arXiv},
      primaryClass={cs.CL},
      url={https://arxiv.org/abs/2304.05302}, 
}

@misc{wei2022finetunedlanguagemodelszeroshot,
      title={Finetuned Language Models Are Zero-Shot Learners}, 
      author={Jason Wei and Maarten Bosma and Vincent Y. Zhao and Kelvin Guu and Adams Wei Yu and Brian Lester and Nan Du and Andrew M. Dai and Quoc V. Le},
      year={2022},
      eprint={2109.01652},
      archivePrefix={arXiv},
      primaryClass={cs.CL},
      url={https://arxiv.org/abs/2109.01652}, 
}

@inproceedings{wang2022super,
  title={Super-naturalinstructions: Generalization via declarative instructions on 1600+ nlp tasks},
  author={Wang, Yizhong and Mishra, Swaroop and Alipoormolabashi, Pegah and Kordi, Yeganeh and Mirzaei, Amirreza and Naik, Atharva and Ashok, Arjun and Dhanasekaran, Arut Selvan and Arunkumar, Anjana and Stap, David and others},
  booktitle={Proceedings of the 2022 conference on empirical methods in natural language processing},
  pages={5085--5109},
  year={2022}
}

@inproceedings{wang2023self,
  title={Self-instruct: Aligning language models with self-generated instructions},
  author={Wang, Yizhong and Kordi, Yeganeh and Mishra, Swaroop and Liu, Alisa and Smith, Noah A and Khashabi, Daniel and Hajishirzi, Hannaneh},
  booktitle={Proceedings of the 61st annual meeting of the association for computational linguistics (volume 1: long papers)},
  pages={13484--13508},
  year={2023}
}

@article{schick2023toolformer,
  title={Toolformer: Language models can teach themselves to use tools},
  author={Schick, Timo and Dwivedi-Yu, Jane and Dess{\`\i}, Roberto and Raileanu, Roberta and Lomeli, Maria and Hambro, Eric and Zettlemoyer, Luke and Cancedda, Nicola and Scialom, Thomas},
  journal={Advances in neural information processing systems},
  volume={36},
  pages={68539--68551},
  year={2023}
}

@article{patil2024gorilla,
  title={Gorilla: Large language model connected with massive apis},
  author={Patil, Shishir G and Zhang, Tianjun and Wang, Xin and Gonzalez, Joseph E},
  journal={Advances in Neural Information Processing Systems},
  volume={37},
  pages={126544--126565},
  year={2024}
}

@article{zheng2023judging,
  title={Judging llm-as-a-judge with mt-bench and chatbot arena},
  author={Zheng, Lianmin and Chiang, Wei-Lin and Sheng, Ying and Zhuang, Siyuan and Wu, Zhanghao and Zhuang, Yonghao and Lin, Zi and Li, Zhuohan and Li, Dacheng and Xing, Eric and others},
  journal={Advances in neural information processing systems},
  volume={36},
  pages={46595--46623},
  year={2023}
}

@inproceedings{liu2023g,
  title={G-eval: NLG evaluation using gpt-4 with better human alignment},
  author={Liu, Yang and Iter, Dan and Xu, Yichong and Wang, Shuohang and Xu, Ruochen and Zhu, Chenguang},
  booktitle={Proceedings of the 2023 conference on empirical methods in natural language processing},
  pages={2511--2522},
  year={2023}
}

@misc{austin2021program,
      title={Program Synthesis with Large Language Models}, 
      author={Jacob Austin and Augustus Odena and Maxwell Nye and Maarten Bosma and Henryk Michalewski and David Dohan and Ellen Jiang and Carrie Cai and Michael Terry and Quoc Le and Charles Sutton},
      year={2021},
      eprint={2108.07732},
      archivePrefix={arXiv},
      primaryClass={cs.PL},
      url={https://arxiv.org/abs/2108.07732}, 
}

@misc{chen2021evaluating,
      title={Evaluating Large Language Models Trained on Code}, 
      author={Mark Chen and Jerry Tworek and Heewoo Jun and Qiming Yuan and Henrique Ponde de Oliveira Pinto and Jared Kaplan and Harri Edwards and Yuri Burda and Nicholas Joseph and Greg Brockman and Alex Ray and Raul Puri and Gretchen Krueger and Michael Petrov and Heidy Khlaaf and Girish Sastry and Pamela Mishkin and Brooke Chan and Scott Gray and Nick Ryder and Mikhail Pavlov and Alethea Power and Lukasz Kaiser and Mohammad Bavarian and Clemens Winter and Philippe Tillet and Felipe Petroski Such and Dave Cummings and Matthias Plappert and Fotios Chantzis and Elizabeth Barnes and Ariel Herbert-Voss and William Hebgen Guss and Alex Nichol and Alex Paino and Nikolas Tezak and Jie Tang and Igor Babuschkin and Suchir Balaji and Shantanu Jain and William Saunders and Christopher Hesse and Andrew N. Carr and Jan Leike and Josh Achiam and Vedant Misra and Evan Morikawa and Alec Radford and Matthew Knight and Miles Brundage and Mira Murati and Katie Mayer and Peter Welinder and Bob McGrew and Dario Amodei and Sam McCandlish and Ilya Sutskever and Wojciech Zaremba},
      year={2021},
      eprint={2107.03374},
      archivePrefix={arXiv},
      primaryClass={cs.LG},
      url={https://arxiv.org/abs/2107.03374}, 
}

@misc{dubois2024length,
      title={Length-Controlled AlpacaEval: A Simple Way to Debias Automatic Evaluators}, 
      author={Yann Dubois and Balázs Galambosi and Percy Liang and Tatsunori B. Hashimoto},
      year={2025},
      eprint={2404.04475},
      archivePrefix={arXiv},
      primaryClass={cs.LG},
      url={https://arxiv.org/abs/2404.04475}, 
}

@misc{he2024multi,
      title={Multi-IF: Benchmarking LLMs on Multi-Turn and Multilingual Instructions Following}, 
      author={Yun He and Di Jin and Chaoqi Wang and Chloe Bi and Karishma Mandyam and Hejia Zhang and Chen Zhu and Ning Li and Tengyu Xu and Hongjiang Lv and Shruti Bhosale and Chenguang Zhu and Karthik Abinav Sankararaman and Eryk Helenowski and Melanie Kambadur and Aditya Tayade and Hao Ma and Han Fang and Sinong Wang},
      year={2024},
      eprint={2410.15553},
      archivePrefix={arXiv},
      primaryClass={cs.CL},
      url={https://arxiv.org/abs/2410.15553}, 
}

@misc{hendrycks2021measuring,
      title={Measuring Mathematical Problem Solving With the MATH Dataset}, 
      author={Dan Hendrycks and Collin Burns and Saurav Kadavath and Akul Arora and Steven Basart and Eric Tang and Dawn Song and Jacob Steinhardt},
      year={2021},
      eprint={2103.03874},
      archivePrefix={arXiv},
      primaryClass={cs.LG},
      url={https://arxiv.org/abs/2103.03874}, 
}

@misc{pal2024smaug,
      title={Smaug: Fixing Failure Modes of Preference Optimisation with DPO-Positive}, 
      author={Arka Pal and Deep Karkhanis and Samuel Dooley and Manley Roberts and Siddartha Naidu and Colin White},
      year={2024},
      eprint={2402.13228},
      archivePrefix={arXiv},
      primaryClass={cs.CL},
      url={https://arxiv.org/abs/2402.13228}, 
}

@misc{rein2023gpqa,
      title={GPQA: A Graduate-Level Google-Proof Q\&A Benchmark}, 
      author={David Rein and Betty Li Hou and Asa Cooper Stickland and Jackson Petty and Richard Yuanzhe Pang and Julien Dirani and Julian Michael and Samuel R. Bowman},
      year={2023},
      eprint={2311.12022},
      archivePrefix={arXiv},
      primaryClass={cs.AI},
      url={https://arxiv.org/abs/2311.12022}, 
}

@misc{yang2025qwen3,
      title={Qwen3 Technical Report}, 
      author={An Yang and Anfeng Li and Baosong Yang and Beichen Zhang and Binyuan Hui and Bo Zheng and Bowen Yu and Chang Gao and Chengen Huang and Chenxu Lv and Chujie Zheng and Dayiheng Liu and Fan Zhou and Fei Huang and Feng Hu and Hao Ge and Haoran Wei and Huan Lin and Jialong Tang and Jian Yang and Jianhong Tu and Jianwei Zhang and Jianxin Yang and Jiaxi Yang and Jing Zhou and Jingren Zhou and Junyang Lin and Kai Dang and Keqin Bao and Kexin Yang and Le Yu and Lianghao Deng and Mei Li and Mingfeng Xue and Mingze Li and Pei Zhang and Peng Wang and Qin Zhu and Rui Men and Ruize Gao and Shixuan Liu and Shuang Luo and Tianhao Li and Tianyi Tang and Wenbiao Yin and Xingzhang Ren and Xinyu Wang and Xinyu Zhang and Xuancheng Ren and Yang Fan and Yang Su and Yichang Zhang and Yinger Zhang and Yu Wan and Yuqiong Liu and Zekun Wang and Zeyu Cui and Zhenru Zhang and Zhipeng Zhou and Zihan Qiu},
      year={2025},
      eprint={2505.09388},
      archivePrefix={arXiv},
      primaryClass={cs.CL},
      url={https://arxiv.org/abs/2505.09388}, 
}

@misc{zhou2023instruction,
      title={Instruction-Following Evaluation for Large Language Models}, 
      author={Jeffrey Zhou and Tianjian Lu and Swaroop Mishra and Siddhartha Brahma and Sujoy Basu and Yi Luan and Denny Zhou and Le Hou},
      year={2023},
      eprint={2311.07911},
      archivePrefix={arXiv},
      primaryClass={cs.CL},
      url={https://arxiv.org/abs/2311.07911}, 
}

\end{document}